\documentclass[journal]{new-aiaa}
\usepackage{packages}

\title{Multi-fidelity aerodynamic data fusion by autoencoder transfer learning}

\author{Javier Nieto-Centenero \footnote{PhD Candidate, Department of Aerospace Engineering, UC3M.} \footnote{Research Scientist, Theoretical and Computational Aerodynamics Group, Flight Physics Department, INTA.}}
\affil{Universidad Carlos III de Madrid, Avda. Universidad 30, Legan\'es 28911, Spain}
\affil{Spanish National Institute of Aerospace Technology (INTA), Ctra. Ajalvir Km.4 Torrej\'on de Ardoz 28850, Spain}
 
\author{Esther Andrés Pérez \footnote{Senior Research Scientist, Theoretical and Computational Aerodynamics Group, Flight Physics Department, INTA.}}
\affil{Spanish National Institute of Aerospace Technology (INTA), Ctra. Ajalvir Km.4 Torrej\'on de Ardoz 28850, Spain}
\author{Rodrigo Castellanos\footnote{Assistant Professor, Department of Aerospace Engineering, UC3M. Corresponding author: \url{rcastell@ing.uc3m.es}}}
\affil{Universidad Carlos III de Madrid, Avda. Universidad 30, Legan\'es 28911, Spain}

\begin{document}

\maketitle

\begin{abstract}
Accurate aerodynamic prediction often relies on high-fidelity simulations; however, their prohibitive computational costs severely limit their applicability in data-driven modeling. This limitation motivates the development of multi-fidelity strategies that leverage inexpensive Low-Fidelity (LF) information without compromising accuracy. Addressing this challenge, this work presents a multi-fidelity deep learning framework that combines autoencoder-based transfer learning with a newly developed Multi-Split Conformal Prediction (MSCP) strategy to achieve uncertainty-aware aerodynamic data fusion under extreme data scarcity.  The methodology leverages abundant LF data to learn a compact latent physics representation, which acts as a frozen knowledge base for a decoder that is subsequently fine-tuned using scarce HF samples. Tested on surface-pressure distributions for NACA airfoils ($2D$) and a transonic wing ($3D$) databases, the model successfully corrects LF deviations and achieves high-accuracy pressure predictions using minimal HF training data. Furthermore, the MSCP framework produces robust, actionable uncertainty bands with pointwise coverage exceeding 95\%. By combining extreme data efficiency with uncertainty quantification, this work offers a scalable and reliable solution for aerodynamic regression in data-scarce environments.

\end{abstract}

\section*{Nomenclature}
{\renewcommand\arraystretch{1.05}
\noindent\begin{longtable*}{@{}l @{\quad=\quad} p{0.72\textwidth}@{}}
\multicolumn{2}{@{}l}{\textbf{Latin Symbols}} \\
$B$ & Number of splits in MSCP \\
$b/2$ & Semi-wing span \\
$C_D$ & Drag coefficient \\
$C_L$ & Lift coefficient \\
$C_p$ & Pressure coefficient \\
$C_{\delta}$ & Conformal Prediction region \\
$c$ & Airfoil chord length \\
$D$ & Dimensionality of the physical space \\
$d$ & Dimensionality of the latent space \\
$E$ & Number of training epochs \\
$f_{\theta}$ & Encoder function \\
$g_{\phi}$ & Decoder function \\
$h_{\psi}$ & Adaptation layer function \\
$\hat{k}$ & Critical quantile index \\
$k_s$ & Critical quantile for conformal prediction \\
$\mathbf{M}$ & Predictive model \\
$M$ & Maximum camber \\
$N$ & Total number of available snapshots \\
$n_z$ & Load factor \\
$P$ & Position of maximum camber \\
$\mathcal{Q}$ & Set of data samples \\
$r$ &  Normalized residual \\
$\mathbf{R}$ & Uncertainty band radius \\
$Re$ & Reynolds number \\
$S$ & Non-conformity score \\
$\mathbf{s}$ & Spatial modulation vector \\
$\mathbf{x}$ & Input vector \\
$\hat{\mathbf{x}}$ & Intermediate high-fidelity basis \\
$\textit{XX}$ & Maximum thickness \\
$\mathbf{y}$ & Output vector (Ground truth) \\
$\mathbf{z}$ & Latent vector \\[1em]

\multicolumn{2}{@{}l}{\textbf{Greek Symbols}} \\
$\alpha$ & Angle of attack \\
$\gamma$ & Latent space coordinates \\
$\delta$ & Significance level \\
$\epsilon$ & Residual error \\
$\eta$ & Spanwise section coordinate \\
$\theta$ & Encoder trainable parameters \\
$\phi$ & Decoder trainable parameters \\
$\psi$ & Adaptation layer trainable parameters \\[1em]

\multicolumn{2}{@{}l}{\textbf{Subscripts, Superscripts and Modifiers}} \\
$\tilde{(\cdot)}$ & Model prediction \\
$(\cdot)^*$ & Median-aggregated quantity \\
$_{(\cdot)^{(b)}}$ & MSCP split iteration index \\
$_{(\cdot)_{cal}}$ & Calibration set \\
$_{(\cdot)_{HF}}$ & High-Fidelity \\
$_{(\cdot)_{LF}}$ & Low-Fidelity \\
$_{(\cdot)_{train}}$ & Training set \\
\end{longtable*}}

\twocolumn 

\section{Introduction}
Modern aerospace design faces a fundamental tension between the demand for predictive accuracy and the constraints of computational resources \cite{slotnick2014cfd}. High-Fidelity (HF) Computational Fluid Dynamics (CFD) methods remain indispensable for capturing complex nonlinear phenomena, such as shock waves and boundary-layer separation. Still, even with recent algorithmic and hardware advancements \cite{gasparino2024sod2d, guerrero2025python}, their expense limits their use in large design studies and iterative optimization. 
Conversely, Low-Fidelity (LF) approaches, such as potential-flow or Euler-based models, are computationally efficient, but they generally lack the viscous, transition, and separation physics required for reliable prediction in complex aerodynamic regimes.
This imbalance has motivated the emergence of Multi-Fidelity (MF) data fusion, whose goal is to combine the abundance of \textit{rapid} LF data with the physical veracity of sparse HF samples. By learning the underlying correlations among fidelities, MF surrogate models achieve predictive accuracy comparable to realistic simulations at a fraction of the cost. This capability is crucial for enabling advanced optimization workflows and reducing time-to-market in the aeronautical industry.

Historically, statistical regression techniques, particularly Co-Kriging (also known as Multi-Fidelity Gaussian Process Regression), have served as the benchmark for MF \cite{kennedy2000predicting, forrester2007multi}. Within the aerospace domain, these methodologies have been extensively explored, employing both hierarchical frameworks \cite{forrester2006optimization, han2010new, kuya2011multifidelity, han2012hierarchical, arenzana2021multi}, where LF data informs the trend of HF models, and non-hierarchical approaches \cite{lam2015multifidelity,feldstein2020multifidelity}. In the latter case, the framework allows assigning specific confidence levels to each information source based on prior experience or empirical validation. While several studies have attempted to extend these methodologies to predict full flow fields \cite{anhichem2022multifidelity, Nieto-Centenero2023, nieto2024multifidelity}, Gaussian Processes (GP) remain predominantly effective for scalar Quantities of Interest, such as lift ($C_L$) and drag ($C_D$) coefficients. When applied to high-dimensional outputs, these methods encounter severe scalability limitations due to the cubic computational complexity of covariance matrix inversion.

To address the challenge of high-dimensional field reconstruction, the Gappy Proper Orthogonal Decomposition (GPOD) \cite{everson1995karhunen} was introduced as a method to recover complete data vectors from incomplete or \textit{gappy} inputs. This technique was adapted for  MF aerodynamic applications, enabling the reconstruction of full flow fields from sparse sensor measurements or limited HF samples \cite{mifsud2019fusing, bui2004aerodynamic}. Building upon this framework, subsequent developments have sought to incorporate uncertainty quantification capabilities into the reconstructed solutions \cite{bertram2021fusing, renganathan2020aerodynamic}. However, the inherent linearity of the underlying basis restricts its ability to accurately represent highly nonlinear flow features with sharp discontinuities, such as moving shock waves.

To circumvent the scalability limitations of GP and the linearity constraints of POD, the field has shifted towards Deep Learning (DL). Techniques include coupling dimensionality reduction with recurrent networks for time-dependent problems \cite{Conti2024}, exploring multi-level neural network architectures to learn nonlinear correlations, or integrating experimental and numerical data directly into the loss function via adaptive weighting \cite{guo2022multi, li2022deep}. However, to address the inherent scarcity of HF data, Transfer Learning has emerged as the dominant paradigm, enabling the fine-tuning of networks pre-trained on abundant LF data to optimize aerodynamic shapes \cite{wu2024efficient}, extract domain-invariant features via adaptation techniques \cite{Kou2022}, predict complex physical properties \cite{wong2025inductive}, or correct systematic errors by fusing simulations with sparse HF information \cite{barklage2025fusing}. In the context of full flow-field prediction, where high dimensionality poses a significant challenge, methodologies have evolved from manifold alignment techniques for inconsistent meshes \cite{perron2021multi} to more sophisticated Autoencoder (AE) frameworks. Current approaches encompass Convolutional (CAE) \cite{mondal2021multi} and Variational Autoencoders (VAE) \cite{cheng2024bi}, architectures leveraging residual connections and super-resolution to enhance efficiency \cite{partin2023multifidelity, shen2024application}, and sequential training algorithms designed to construct hierarchical and interpretable latent spaces \cite{saetta4898782autoencoders, najafi2025introducing}.

Finally, within the context of robust aerospace design, predictive accuracy must be accompanied by a reliability measure, a critical requirement for managing epistemic errors \cite{slotnick2014cfd}. To address this challenge, recent literature has explored advanced probabilistic approaches, such as Deep Ensembles \cite{saetta2024uncertainty}, Bayesian Neural Networks \cite{vaiuso2024multi}, or GP \cite{lee2025large}, which aim to characterize model uncertainty. However, as these methods often rely on strict distributional assumptions or entail high computational costs, Conformal Prediction \cite{angelopoulos2023conformal} has emerged as a robust alternative providing distribution-free statistical coverage guarantees for finite samples. As detailed in \cite{gopakumar2024uncertainty}, this technique facilitates the construction of rigorous uncertainty bands, a capability that has recently been utilized to quantify uncertainty in complex regression tasks such as assessing the reliability of dynamics systems \cite{liang2024conformal} or securing semi-autonomous aircraft navigation during landing maneuvers \cite{vilfroy2024conformal}.

Despite recent advances, existing MF surrogate modeling techniques still struggle to combine high predictive accuracy with reliable, data-efficient uncertainty quantification in the low-data regime. To address this gap, this paper presents a comprehensive deep learning framework for aerodynamic data fusion that integrates a Multi-Fidelity Autoencoder (MFAE) architecture with a data-driven uncertainty quantification strategy. The methodology leverages abundant LF data to pre-train the encoder, extracting a compact latent representation of the underlying physics that serves as a frozen knowledge base. The decoder is then fine-tuned using scarce HF samples, while a dedicated adaptation layer corrects LF-to-HF biases and bridges dimensionality differences. 
To assess uncertainty, we introduce a Multi-Split Conformal Prediction (MSCP) approach tailored for data-scarce regimes. The central contribution of this strategy lies in its data efficiency and robustness. Unlike standard conformal protocols that permanently reserve a dedicated fraction of the dataset for calibration, thereby reducing the information available for learning, MSCP iteratively resamples the HF data, enabling full utilization of all available HF samples during the final fine-tuning stage. By aggregating calibration statistics across multiple partitions, the method mitigates the high sensitivity to a single random split and yields consistent, physically meaningful uncertainty bounds. The same resampling loop provides a statistically grounded estimate of the optimal training epochs required to prevent overfitting, thereby eliminating the need for a separate validation set. Unlike standard applications that seek strict nominal coverage guarantees, which often result in excessively conservative intervals of limited value for decision-making, this work prioritizes pointwise coverage to generate narrower, actionable uncertainty bands, thereby balancing statistical rigor with practical engineering utility.
The proposed framework is validated on two aerodynamic datasets of increasing complexity, demonstrating high predictive accuracy, strong data efficiency, and reliable uncertainty quantification across both 2D airfoils and a 3D transonic wing.

\section{PROPOSED FRAMEWORK AND METHODS}
This section presents the central elements of the proposed methodological framework. The MF surrogate model, implemented via an autoencoder architecture, is initially described as the predictive core. The following subsection introduces the uncertainty quantification strategy used to evaluate model reliability. Finally, the integration of these components into a cohesive, automated computational pipeline is detailed, providing a comprehensive view of the end-to-end workflow.

\subsection{Multi-Fidelity Surrogate Modeling via Autoencoder Architecture}
In the context of deep learning, autoencoders constitute a key architecture for unsupervised representation learning. Their design is based on the premise that a neural network can learn efficient, low-dimensional encoding of input data while simultaneously preserving the essential information required for accurate reconstruction \cite{hinton2006reducing}. This ability to capture and represent latent features is particularly relevant in domains of high complexity and dimensionality.

An AE architecture is organized into two sequential modules: an encoder and a decoder. The input data is denoted as a vector $\mathbf{x}$ in the data space $\mathcal{X} \subset \mathbb{R}^D$. The encoder maps this input to a latent vector $\mathbf{z}$ in a lower-dimensional latent space $\mathcal{Z} \subset \mathbb{R}^d$ (where $d \ll D$), while the decoder performs the inverse operation. Mathematically, this process is expressed as the composition of two functions: the encoder $f_\theta: \mathcal{X} \rightarrow \mathcal{Z}$, and the decoder $g_\phi: \mathcal{Z} \rightarrow \mathcal{X}$. The subscripts $\theta$ and $\phi$ represent the trainable parameters (weights and biases) of the respective networks. The data is compressed into the latent space in the bottleneck, resulting in the reconstructed output defined as $\tilde{\mathbf{x}} = (g_\phi \circ f_\theta)(\mathbf{x})$. This model is trained to minimize a reconstruction loss function $\mathcal{L}_{rec}$, typically the Mean Squared Error (MSE), to ensure the output matches the input data.

\begin{figure*}[t]
    \centering
    \includegraphics[width = 0.99\textwidth]{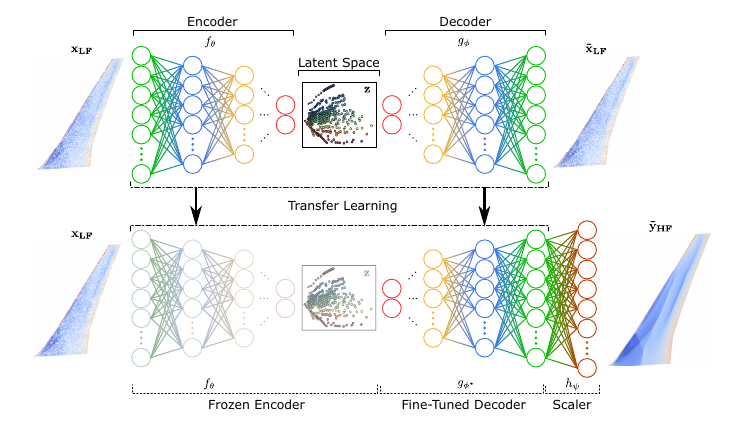}
    \caption{Overview of the training workflow used in the proposed multi-fidelity autoencoder, illustrating the interaction between low- and high-fidelity data streams.}
    \label{MF_autoencoder}
\end{figure*}

The methodology is based on an MFAE, as shown in \autoref{MF_autoencoder}, designed to address the typical asymmetry in simulation data availability: abundant LF data versus scarce HF data. This approach leverages LF data to learn fundamental system representations, which are subsequently refined using the more precise and costly HF data. To achieve this, the training is conducted in two distinct phases that progressively build the final architecture.

In the first phase, the model is established using a conventional autoencoder scheme trained exclusively on LF data. Let $\mathbf{x}_{LF} \in \mathbb{R}^{D_{LF}}$ be the input vector. The encoder $f_\theta: \mathbb{R}^{D_{LF}} \rightarrow \mathbb{R}^d$ maps the LF input to the latent space, and the decoder $g_\phi: \mathbb{R}^d \rightarrow \mathbb{R}^{D_{LF}}$ reconstructs it back to the original LF dimension. By minimizing the MSE between the input ($\mathbf{x}_{LF}$) and its reconstruction ($\tilde{\mathbf{x}}_{LF}$), the imposed dimensionality bottleneck constrains the encoder to extract the most relevant features and identify the principal modes governing system behavior. The resulting encoder weights $\theta$ constitute a knowledge base preserved for the subsequent phase.

In the second phase, the model performs the transfer to the high-fidelity domain. The encoder parameters $\theta$ are frozen under the assumption that the underlying physics remain consistent across fidelities, thereby preserving the physically-consistent latent manifold established during the LF pre-training. In regimes of extreme HF data scarcity, updating the encoder parameters risks overfitting and distorting these learned global flow structures. Furthermore, keeping the latent space unaltered avoids introducing additional variance across the different data splits, thereby supporting a more consistent and stable calibration within the MSCP framework. Consequently, the input $\mathbf{x_{LF}} \in \mathbb{R}^{D_{LF}}$ is mapped to the latent space by the pre-trained encoder. Optimization focuses on adapting the decoder and training the up-scaler. The decoder is fine-tuned using the scarce HF data (updating parameters to $\phi^*$), generating a reconstruction $\hat{\mathbf{x}} \in \mathbb{R}^{D_{LF}}$ from the latent representation. This intermediate output $\hat{\mathbf{x}}$ acts as a high-fidelity basis in the low-fidelity dimension. Subsequently, to bridge the dimensionality gap, an up-scaler $h_\psi: \mathbb{R}^{D_{LF}} \rightarrow \mathbb{R}^{D_{HF}}$, parametrized by $\psi$, is introduced. Unlike the pre-trained components, the parameters $\psi$ are initialized and trained from scratch to map and refine the decoder's basis $\hat{\mathbf{x}}$ to the final HF dimensionality, incorporating details not present in the LF data. The resulting MF surrogate model is thus defined as a nested composition $\mathbf{M}$, where the final HF prediction $\tilde{\mathbf{y}}_{HF} \in \mathbb{R}^{D_{HF}}$ is obtained from the LF input $\mathbf{x_{LF}} \in \mathbb{R}^{D_{LF}}$ by:
$$\tilde{\mathbf{y}}_{HF} = \mathbf{M}(\mathbf{x_{LF}}) = (h_\psi \circ g_{\phi^*} \circ f_\theta)(\mathbf{x_{LF}})$$

In summary, the MFAE facilitates the transfer of physical knowledge by capturing global structures from abundant LF data. These representations are subsequently refined through the fine-tuned decoder and the up-scaler, which incorporate specific high-fidelity details. By treating the encoder as a fixed feature extractor, the decoder and up-scaler can focus entirely on resolving localized multi-fidelity discrepancies without compromising the overall physical stability. This approach optimizes training efficiency and ensures model robustness, even in regimes of severe HF data scarcity.

\subsection{Enhanced Conformal Prediction via Multi-Split Calibration}

Standard regression models typically provide point predictions without quantifying the confidence of the estimation. To address this, CP offers a framework to construct uncertainty regions with statistical coverage guarantees, relying on the concept of exchangeability rather than strict distributional assumptions \cite{angelopoulos2023conformal}. Unlike Bayesian methods that depend on priors, CP uses past errors from a calibration set to learn the appropriate uncertainty magnitude. However, in aerodynamic flows, the prediction error is not uniform, but tends to be concentrated in regions with complex physics, such as shock waves or the leading edge. Therefore, a uniform uncertainty band would be inefficient. This work implements a modulated CP strategy, following the approach for multivariate functional data proposed in \cite{diquigiovanni2022conformal}. The core idea is to decouple the uncertainty into two components: a spatial modulation function ($s_j$) that captures the local shape of the error variability, and a global scalar ($k_s$) that adjusts the size of the prediction region, $\mathbf{C}_{\delta}(\mathbf{x})$, to satisfy the desired coverage level $\mathbb{P}(\mathbf{y} \in \mathbf{C}_{\delta}(\mathbf{x})) \geq 1-\delta$.

The construction of these adaptive bands begins by quantifying the complexity of predicting each specific point on the wing surface. A dimension-specific modulation factor $s_j \in \mathbb{R}^+$ is calculated for each mesh node $j \in \{1, \dots, D\}$. This factor locally scales the residuals to handle heteroscedasticity. It is defined as the sample standard deviation of the residuals computed over the training set:
\begin{equation}
    s_j = \sqrt{\mathbb{V}\text{ar}\left( \left\{ \epsilon_{i,j} \right\}_{i \in \mathcal{Q}_{train}} \right)}, \quad \text{where } \epsilon_{i,j} = y_{i,j} - M_j(\mathbf{x}_i) .
    \label{eq:scaling_factor}
\end{equation}

Physically, $s_j$ acts as a data-driven map of the expected error intensity. We utilize this map to normalize the residuals of the calibration set $\mathcal{Q}_{cal}$, obtaining a dimensionless error metric $r_{i,j} = |\epsilon_{i,j}| / s_j$. This step ensures that high-error regions do not dominate the calibration solely because of their magnitude, but rather contribute proportionally to their local variability.

Subsequently, these normalized residuals are spatially aggregated to determine a single global non-conformity score $S_i$ for each snapshot $i$, representing the overall prediction misfit across the entire flow field. This study evaluates two distinct aggregation metrics: the $L_{\infty}$ norm, which tracks the maximum local error, and the Root Mean Squared Residual (RMSR), which measures the grid-wide average quadratic deviation. These scores are defined as:
\begin{equation}
    \begin{aligned}
    \text{$L_{\infty}$ norm:} \quad 
      S_i &= \max_{j = 1,\dots,D} r_{i,j}, \\[4pt]
    \text{RMSR:} \quad
      S_i &= \sqrt{\frac{1}{D} \sum_{j=1}^{D} r_{i,j}^2}.
    \end{aligned}
\end{equation}
With the set of scores $\{S_i\}_{i=1}^{n_{cal}}$ obtained, the calibration procedure identifies the critical quantile $k_s$. This threshold corresponds to the $\hat{k}$-th smallest value among the scores, where $\hat{k} = \min \left( \lceil (n_{cal} + 1)(1-\delta) \rceil, \, n_{cal} \right)$, thereby guaranteeing a marginal coverage of at least $1-\delta$ for the selected non-conformity score under exchangeability.

Finally, the uncertainty region is mapped back to the physical domain. For engineering visualization and direct structural comparison, the prediction intervals for both metrics are plotted as a single spatial vector band, defined as:
\begin{equation}
    \mathbf{C}^{\text{band}}_{\delta}(\mathbf{x}_*) = \left[ \mathbf{M}(\mathbf{x}_*) - \mathbf{R}, \quad \mathbf{M}(\mathbf{x}_*) + \mathbf{R} \right],
\end{equation}
where $\mathbf{M}(\mathbf{x}_*) \in \mathbb{R}^D$ is the full predicted flow field vector, and $\mathbf{R} = k_s \mathbf{s} \in \mathbb{R}^D$ is the vector of adaptive band half-widths.

Crucially, the statistical coverage guarantee applies differently to this plotted band depending on the chosen metric. For the $L_{\infty}$ norm, $\mathbf{C}^{\text{band}}_{\delta}(\mathbf{x}_*)$ matches the exact theoretical conformal region, ensuring strict simultaneous coverage. For the RMSR metric, however, the exact mathematical guarantee applies strictly to an ellipsoidal confidence region in $\mathbb{R}^D$. Consequently, presenting the RMSR bounds as a hyperrectangular band is a visualization choice representing the local scale of the calibrated error; it does not guarantee simultaneous coverage, meaning the continuous flow field may locally cross the plotted boundaries as long as the global grid-wide quadratic error remains confined within the theoretical ellipsoid.

Instead of traditional parametric confidence intervals (e.g., those derived from Kriging or Gaussian Processes), this formulation yields an empirical, distribution-free prediction region. Under this framework, uncertainty adapts locally, widening in regions with high training variability and shrinking in regions where the model is more confident. This yields a physically interpretable uncertainty quantification that adapts to the local flow features while maintaining global statistical rigor.

\subsubsection*{Multi-Split Conformal Prediction for Data-Scarce Environments}

While CP provides a rigorous framework for uncertainty quantification, its dependence on a single, static partition of data into training and calibration sets introduces significant limitations in data-scarce domains. In regimes with limited sample sizes, dedicating a substantial fraction of data solely to calibration can degrade the model's predictive capability. Furthermore, the resulting prediction bands exhibit high sensitivity to the specific random partition employed; a single unrepresentative split may yield misleading uncertainty estimates. To mitigate these drawbacks, we propose a Multi-Split Conformal Prediction (MSCP) methodology. This approach uses an ensemble procedure to generate stable and reliable prediction bands while enabling the final predictive model to be trained on the entire available HF data.

Before delving into the algorithm, it is essential to address a specific challenge inherent in deep learning in data-scarce regimes. Although the model quickly learns the global characteristics of the flow, thereby minimizing validation error, its generalization ability eventually stagnates due to the limited dataset size. Despite this plateau, continuous training forces the model to reduce the loss further. This minimization is achieved not by learning generalizable features, but by memorizing specific high-frequency details intrinsic to training samples \cite{spectral_bias}. This results in training residuals that are artificially low, failing to reflect the actual uncertainty observed on unseen data. Deriving modulation functions solely from these residuals leads to numerical instabilities and excessively wide prediction intervals during calibration. To address this robustly, our MSCP implementation modifies the standard protocol by computing the modulation vector using the calibration residuals of each split. Although this relaxes the strict exchangeability assumption, the subsequent robust aggregation across multiple splits effectively compensates for this, aligning the uncertainty quantification with the empirical generalization error and promoting engineering utility.

The MSCP calibration process consists of the following steps:
\begin{enumerate}
    \item \textbf{Iterative Re-sampling and Calibration:} The procedure performs $B$ iterations. In each iteration $b \in \{1, \dots, B\}$, the available HF dataset is randomly partitioned into a temporary training subset $\mathcal{Q}_{train}^{(b)}$ and a calibration subset $\mathcal{Q}_{cal}^{(b)}$. An auxiliary predictive model, $\mathbf{M}^{(b)}$, is trained on $\mathcal{Q}_{train}^{(b)}$. Subsequently, the standard CP procedure is adapted: dimension-specific modulation functions $\mathbf{s}^{(b)}$ are computed directly from the residuals on $\mathcal{Q}_{cal}^{(b)}$ to address the generalization gap described above. Then, a critical quantile score $k_s^{(b)}$ is determined from non-conformity scores calculated on the same $\mathcal{Q}_{cal}^{(b)}$. This yields a candidate calibration radius vector for the current split, defined as $\mathbf{R}^{(b)} = k_s^{(b)} \mathbf{s}^{(b)}$.

    \item \textbf{Collection of Calibration Margins:} After $B$ iterations, a set of $B$ candidate radial vectors is obtained: $\{\mathbf{R}^{(b)}\}_{b=1}^B$. Each vector $\mathbf{R}^{(b)}$ represents the uncertainty margin for that specific data partition.

    \item \textbf{Robust Statistic Aggregation:} The final calibration margin is derived by robustly aggregating the candidate radii. For each response dimension $j$, the component-wise median of the collection $\{R_j^{(1)}, \dots, R_j^{(B)}\}$ is computed. Formally, the final stable radius is defined as $R^*_j = \text{median}(\{R_j^{(b)}\}_{b=1}^B)$. The median is deliberately selected over alternative aggregation metrics for its robustness against outliers; it effectively mitigates extreme values derived from unrepresentative data partitions, ensuring that the final uncertainty estimate reflects the central tendency of model performance rather than isolated anomalies.
\end{enumerate}

This strategy is preferred over structured alternatives like CV+ \cite{barber2021predictive}. While CV+ offers rigorous coverage guarantees, its rigid partition structure is sensitive to fold assignments in data-scarce regimes, where unrepresentative folds can skew estimation. MSCP mitigates this risk via randomized resampling and median aggregation. Although this approach sacrifices strict theoretical validity, priority is placed on stability to produce informative uncertainty margins, ensuring the resulting bands are actionable.

Significantly, this framework circumvents a primary bottleneck in practical applications: the data efficiency penalty associated with splitting. In contrast to standard protocols that reserve a fraction of the dataset for calibration, our methodology employs resampling to define uncertainty bounds, thereby enabling the final model to be trained on the entire set of HF samples. This is particularly important in engineering contexts where data generation incurs high computational cost. It is worth noting, however, that this statistical robustness introduces an internal computational trade-off. Since the fine-tuning and calibration loop must be repeated for each of the $B$ random splits, the overall training time scales approximately linearly with the number of iterations relative to a single-split approach. Nevertheless, this additional computational cost is justified by the considerable improvements in the stability and reliability of the resulting uncertainty intervals.

The present study constitutes the first application-specific assessment of the proposed MSCP variant in an aerodynamic surrogate-modeling context. In this framework, the calibration performed within each split follows the standard multivariate CP formulation described above and is implemented from the referenced methodology. The novelty introduced here lies in the repeated resampling, split-wise recalibration, and robust aggregation of the resulting margins, with the aim of improving stability in severe HF data-scarce regimes. Accordingly, this work does not claim a new strict theoretical coverage guarantee for the aggregated MSCP procedure beyond those associated with standard CP at the split level. Instead, the practical validity of the method is assessed empirically on the two case studies considered in this paper, described in \autoref{test_cases}.

\begin{figure*}[h!]
    \centering
    \includegraphics[width = 0.88\textwidth]{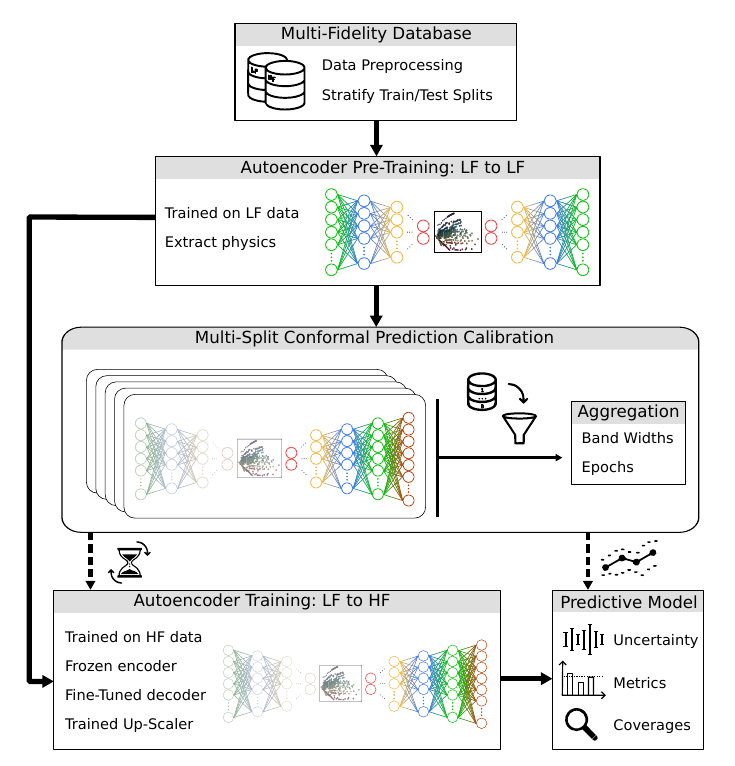}
    \caption{Full methodological pipeline. The methodology integrates LF pre-training, MSCP for uncertainty calibration, and fine-tuning for HF predictions.}
    \label{pipeline}
\end{figure*}

\subsection{Methodological Pipeline}
To address the challenge of predicting aerodynamic behavior from MF data while reliably quantifying uncertainty, we designed a comprehensive pipeline that integrates transfer learning with the proposed MSCP framework, as illustrated in \autoref{pipeline}. The workflow begins with a systematic data structuring phase. The available HF dataset is partitioned into a training set and a Test set via stratified sampling across design variables, ensuring that both subsets are statistically representative of the entire operational envelope. Crucially, to avoid data leakage, the LF counterparts of the HF test cases are strictly excluded from the training set, guaranteeing that the model remains entirely unexposed to the test geometries during the development stages.

The predictive core is founded upon a deep autoencoder, initially trained with comprehensive LF data to extract representations of fundamental features. The architecture imposes a dimensionality bottleneck, setting the latent space dimension equal to that of the physical design space. This choice, supported by prior work \cite{frances2024toward}, aims to enhance interpretability by forcing the model to map the latent variables to the design parameters governing the airflow. By minimizing the reconstruction error of LF inputs, the encoder learns to compress the constitutive physics and global flow structures into this compact manifold. Simultaneously, the decoder acquires a robust initialization, providing an informed starting point for the subsequent transfer learning stage.

Following the configuration of the transfer learning architecture, the MSCP protocol is executed to calibrate uncertainty and determine optimal training epochs. The pre-trained model is cloned $B$ times, and for each replicate $b$, the HF training set is randomly partitioned into temporary training ($\mathcal{Q}_{train}^{(b)}$) and calibration ($\mathcal{Q}_{cal}^{(b)}$) subsets. A fine-tuning process is then executed using an Early-Stopping criterion. In each iteration, the temporary calibration subset serves as a validation set to monitor generalization error, halting the training process after 100 epochs without improvement to prevent overfitting. During this procedure, two key metrics are recorded: the calibrated uncertainty margins $R^{(b)}$ and the specific number of epochs $E^{(b)}$ required to reach convergence for that particular data partition.

The final step leverages these statistics to maximize data utility. By computing the median of the recorded epochs, $E^* = \text{median}(\{E^{(b)}\}_{b=1}^B)$, a stable, data-driven stopping criterion is established. Consequently, the final predictive model is fine-tuned using the entirety of the available HF training data, terminating the process strictly at epoch $E^*$. This strategy eliminates the need for a separate validation set during the final training phase, thereby maximizing the information available for model optimization while maintaining a robust safeguard against overfitting. The output is a predictor coupled with a stable uncertainty band $\mathbf{R}^*$, which is finally evaluated on the unseen Test set to verify performance metrics and empirical coverage.

\section{REPRESENTATIVE TEST CASES} \label{test_cases}
To thoroughly validate the proposed methodology, this investigation employs two distinct datasets characterized by increasing complexity. The first dataset comprises two-dimensional airfoil geometries, while the second extends the analysis to three-dimensional wing configurations. In both scenarios, the primary objective is to predict the Pressure Coefficient ($C_p$) distribution over the respective aerodynamic surfaces. This quantity was selected because it is the common paired LF/HF observable consistently available across both benchmark datasets and because it directly reflects the strongest nonlinear aerodynamic features of interest in the present study. Accordingly, the validation reported here should be interpreted as a $C_p$-focused assessment of the proposed methodology.

\subsection{Airfoils Database}
The first validation case utilizes the \textit{Airfoils database}, which comprises $C_p$ vectors obtained from CFD simulations of 443 NACA four-digit airfoils, all evaluated at a fixed angle of attack of $\alpha = 10^\circ$. The geometric space is parameterized by three variables: maximum camber ($M \in [2, 8]\%$ of the chord), location of maximum camber ($P \in [5, 8]$ tenths of the chord), and maximum thickness ($XX \in [8, 20]\%$ of the chord). These parameters define a diverse geometric design space suitable for evaluating the framework's robustness.

\subsubsection{High-Fidelity Dataset}
The HF ground truth consists of Reynolds-Averaged Navier-Stokes (RANS) simulations of turbulent flow at $Re = 3 \times 10^6$. The raw data is publicly available \cite{schillaci2021_dataset_1, schillaci2021_dataset_2}, with detailed generation protocols described in \cite{schillaci2020}.
While the original simulations provide 1500 points per airfoil, the present study downsamples this resolution to 260 points using a cosine distribution. This sampling strategy clusters nodes near the leading and trailing edges to capture steep pressure gradients, thereby preserving the global fidelity of the pressure distribution while reducing dimensionality for the analysis.

\subsubsection{Low-Fidelity Dataset}

The LF data is generated using the inviscid panel formulation implemented in \texttt{XFOIL} \cite{xfoil}. We stress that, in this study, \texttt{XFOIL} is not used in its viscous/inviscid interaction mode, but as a deliberately low-cost inviscid auxiliary model. Under the selected reference conditions ($Re = 3 \times 10^6$, $\alpha = 10^\circ$, and a broad range of airfoil geometries), this choice introduces a substantial and intentional physical discrepancy relative to the turbulent RANS HF dataset, since the LF model does not represent the viscous, transition, and separation mechanisms embedded in the HF reference. The purpose of this choice is to define a demanding transfer-learning scenario in which the proposed framework must recover HF pressure behavior from an intentionally simplified LF source.

Unlike the fixed resolution of the HF dataset, the LF data is generated with varying discretization densities to evaluate the impact of resolution up-scaling on multi-fidelity modeling errors. This is achieved by defining specific input proportions relative to the original HF point count. For each selected proportion, the LF pressure coefficient vectors are generated by interpolating the source data onto these reduced-size grids, ensuring that each dedicated network instance operates on a consistent, fixed-dimensional input. This systematic reduction allows for an explicit investigation of the framework's ability to reconstruct high-resolution fields from spatially sparse LF information without altering the underlying model architecture.

\begin{figure}[t]
    \centering
    \includegraphics[width = 8.2 cm]{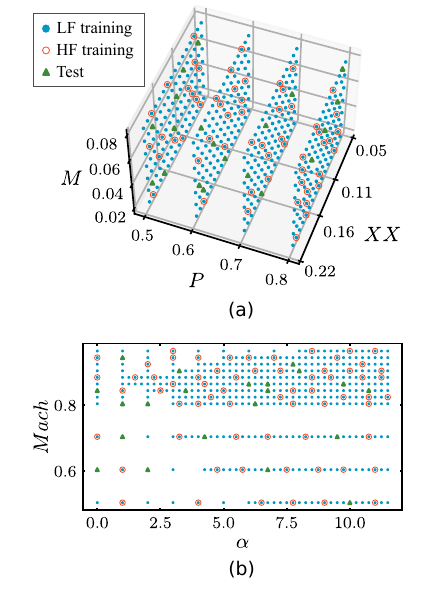}
    \caption{Datasets: (a) \textit{Airfoils database} (NACA) defined by M, P, and XX; (b) \textit{Wing database} (XRF1) under varying Mach number and angle of attack ($\alpha$).} 
    \label{train_test_split}
\end{figure}
\subsubsection{Train-Test Split}
The dataset is partitioned following the sampling logic described in the methodological pipeline. To replicate the scarcity constraints of a realistic setting, the original pool of 443 cases is downsampled to a restricted HF set of 79 snapshots, selected via stratified random sampling to ensure the training set is statistically representative across the entire range of geometric variations.

From the HF subset, 75\% (59 cases) are allocated for training, while the remaining 25\% (20 cases) are reserved as a Test set, as illustrated in \autoref{train_test_split} (a). The available LF training data comprises all original cases excluding those assigned to the Test set in HF, guaranteeing that the latter remains strictly isolated during the learning process.

Additionally, a partition termed the \textit{Complementary} Test Set is established, comprising the 384 cases excluded from the HF training budget. In an operational context, HF simulations are strictly limited to the designated Training and Test sets. However, as this study relies on a fully pre-computed database, HF solutions for these unselected geometries remain available. Consequently, this unutilized data is leveraged to benchmark the model across the entire geometric manifold. This assessment is crucial to verify that the error metrics observed in the Test Set reflect a robust global generalization capability and are not the result of a specific sample selection.

\subsection{Wing Database} \label{Garteur_database}
The wing case study considered in this work is based on a CFD database of the XRF1 research aircraft model. This model, provided by Airbus\texttrademark, was developed as a benchmark configuration to investigate advanced technologies for wide-body, long-range aircraft. The database was generated within the AD/AG60 project of the Group for Aeronautical Research and Technology in Europe (GARTEUR \cite{garteur_web}). Further details about the XRF1 model can be found in \cite{pattinson2013xrf1}.

\subsubsection{High-Fidelity Dataset}
The HF data is obtained from RANS simulations of the XRF1 aircraft configuration performed with the DLR TAU solver \cite{kroll2014dlr}. The dataset comprises $N=435$ flight conditions at a fixed Reynolds number of $Re = 2.5\times 10^7$, spanning Mach numbers in the range $0.5 \leq Mach \leq 0.96$ and angles of attack $0^\circ \leq \alpha \leq 11.5^\circ$. Crucially, all simulations are conducted at a fixed load factor of $n_z=2.5$, corresponding to a loading condition of clear practical relevance in aircraft certification and structural design. In the present work, however, the XRF1 HF database is treated as a steady RANS dataset defined under fixed flight conditions, rather than as a fully coupled aeroelastic benchmark. Accordingly, the purpose of this case study is not to analyze fluid-structure interaction effects, but to assess the proposed surrogate framework on a challenging transonic wing pressure database spanning a broad range of Mach number and angle of attack conditions. These conditions generate complex nonlinear $C_p$ distributions, including strong shock-wave features, that are well-suited for assessing surrogate performance.

For the purpose of this study, the analysis focuses exclusively on the wing's upper surface. This choice avoids numerical artefacts near the engine pylon and focuses on the more challenging prediction of the suction side, as highlighted in \cite{castellanosassessment}. Consequently, each data sample consists of a $C_p$ distribution defined on an unstructured surface mesh of $D = 49{,}574$ nodes.

\subsubsection{Low-Fidelity Dataset}

In the absence of native LF aerodynamic simulations for the XRF1 configuration, a synthetic dataset is constructed directly from the original snapshots. This process employs a dual filtering strategy designed to emulate the information loss characteristic of lower-fidelity models by sequentially reducing modal complexity and grid density. The detailed mathematical derivation is provided in \autoref{Appendix_A}.

The first stage reduces physical precision using Proper Orthogonal Decomposition (POD). This technique decomposes the flow field into an energy-ranked orthogonal basis, enabling a \textit{modal filtering} strategy that preserves only the most dominant information within the data. In this study, the truncation cuts off the spectrum to retain 96\% of the cumulative variance, which corresponds to keeping the first $r^*=21$ modes. This operation functions as a low-pass filter, maintaining the global aerodynamic topology while deliberately discarding high-frequency details. Consequently, sharp nonlinearities, such as shock waves, appear blurred, effectively mimicking the diffusive behavior typical of lower-accuracy numerical solvers.

Subsequently, the spatial resolution is reduced via the Farthest Point Sampling (FPS) algorithm. Unlike random downsampling, FPS utilizes an iterative greedy strategy to select a subset of points that maximizes the uniformity of surface coverage. This method generates a reduced mask of 4,000 nodes, representing approximately 8\% of the original mesh density. 

Accordingly, this wing case should be interpreted as a controlled degradation-recovery benchmark designed to assess the ability of the framework to reconstruct missing high-fidelity structure from systematically filtered inputs. The combination of these two techniques yields a dataset that preserves the essential global aerodynamic topology while significantly reducing both physical precision and spatial resolution. This setting provides a rigorous and well-controlled assessment of the framework’s reconstruction capabilities under controlled information loss.

\subsubsection{Train-Test Split}
The train-test split protocol for this dataset mirrors the strategy used for the \textit{Airfoils database}, as illustrated in \autoref{train_test_split} (b). From the $435$ paired cases, $20\%$ (87 cases) are selected as HF samples. Of these, $75\%$ (65 cases) are used for training, and $25\%$ (22 cases) are held out for testing. A stratified selection process based on Mach number and $\alpha$ is utilized to guarantee representative coverage of the flight envelope. The flight conditions not included in the HF training are grouped into a Complementary Test set. Evaluation on this large group of unseen data serves to verify the robustness of the proposed framework across the complete flight envelope, ensuring that the error metrics reported on the small Test set are representative of the model's global performance.

\section{MFAE ARCHITECTURE AND LATENT SPACE ANALYSIS}
This section details the MFAE architectures and the specific parameters defining the proposed framework for the considered databases. Subsequently, the topology of the learned latent spaces is examined to evaluate the model's ability to capture the underlying flow physics autonomously and to analyze the correlation between the encoded features and the governing simulation parameters.

\subsection{Model Architecture and Training Configuration}
The MFAE architecture for the considered databases is summarized in \autoref{tab:network_arch}. The neural network architectures employed for both databases share a common foundational structure, with fully connected layers where the Rectified Linear Unit (ReLU) serves as the activation function. The training process is governed by the Adam optimizer, a widely used standard algorithm in the machine learning community due to its computational efficiency and robust convergence properties. Despite these shared algorithmic principles, the depth and width of the networks differ significantly to accommodate the specific dimensionality constraints of each physical problem. The final architectures and training hyperparameters were selected through targeted empirical tuning guided by input/output dimensionality, training stability, and prior experience with similar aerodynamic field-regression problems \citep[e.g.][]{frances2024toward,castellanosassessment}, rather than through an exhaustive automated hyperparameter-optimization search.
\begin{table*}
    \caption{\label{tab:network_arch} Model architecture and hyperparameters for Airfoils and Wing databases.}
    \centering
    \renewcommand{\arraystretch}{1.2}
    \begin{tabular}{l c c}
        \toprule
        Database & Airfoils & Wing \\
        \midrule
        Optimizer & \multicolumn{2}{c}{Adam} \\
        Layers Type & \multicolumn{2}{c}{Fully connected} \\
        Activation Type & \multicolumn{2}{c}{ReLU} \\
        Pre-training Learning Rate & $1 \times 10^{-4}$ & $1 \times 10^{-4}$ \\
        Fine-tuning Learning Rate & $1 \times 10^{-4}$ & $1 \times 10^{-5}$ \\
        Input size ($D_{LF}$) & [40, 66, 104, 156, 222, 260] & 4000 \\
        Encoder & 64 - 32 - 16 & 1024 - 512 - 256 - 64 \\
        Latent Space & 3 & 2 \\
        Decoder & 16 - 32 - 16 & 64 - 256 - 512 - 1024 \\
        Up-Scaler & $1.5 \times D_{LF}$ & 8000 \\
        Output size ($D_{HF}$) & 260 & 49574 \\
        MSCP Splits ($B$) & 30 & 30 \\
        \bottomrule
    \end{tabular}
\end{table*}

For the \textit{Airfoils database}, the architecture is designed to support the sensitivity analysis regarding input resolution. Consequently, the size of the input layer ($D_{LF}$) is not fixed; it varies across the experimental configurations, ranging from a coarse discretization of 40 nodes to a fine density of 260 nodes. The encoder progressively compresses this input vector into a three-dimensional latent space, matching the dimensionality of the geometric parameterization. Subsequently, the decoder transforms the compressed latent features back into the physical domain, yielding an HF $C_p$ vector defined over a fixed resolution of 260 components ($D_{HF}$). To accommodate variable input densities in instances of resolution mismatch, the up-scaler module is dynamically sized to $1.5 \times D_{LF}$ neurons. This proportional scaling ensures that the model maintains sufficient plasticity to correct LF biases regardless of the input resolution.

Conversely, the \textit{Wing database} requires a deeper and wider architecture to manage the high-dimensional transonic flow field. The network processes a fixed LF input vector comprising 4,000 nodes ($D_{LF}$), which is compressed into a compact two-dimensional latent representation. Subsequently, the decoder and up-scaler are tasked with reconstructing the complex flow physics onto an HF surface mesh of 49,574 nodes ($D_{HF}$). In this configuration, the up-scaler incorporates a hidden layer of 8,000 neurons, providing the necessary capacity to bridge the substantial resolution gap and recover the high-frequency details lost in the LF source.

Regarding the uncertainty quantification strategy, the framework employs the MSCP procedure configured with $B=30$ splits. In each iterative resampling step, 30\% of the available HF training data is randomly partitioned into a temporary calibration subset, $Q_{cal}^{(b)}$. As detailed in the methodology, this subset assumes a dual critical role: it serves as the reference data for calibrating uncertainty bounds and functions as a validation set to monitor the generalization error during the fine-tuning phase, facilitating the precise selection of the stopping epoch.

Finally, to evaluate the practical feasibility of the proposed framework in engineering workflows, a computational scaling analysis was performed on a dedicated workstation equipped with an Intel Core i9-10900X CPU (3.70~GHz), 31~GB of RAM, and a single NVIDIA RTX A5000 GPU (24~GB VRAM). The core neural network operations were accelerated by leveraging the GPU. This analysis reveals that expanding from the 2D airfoil to the highly dense 3D wing configuration increases the number of trainable model parameters by more than three orders of magnitude, from $1.23 \times 10^5$ to $4.38 \times 10^8$. Despite this architectural expansion, the computational overhead scales moderately and is restricted entirely to the offline phase; the total offline budget, comprising Phase 1 pre-training, MSCP calibration, and Phase 2 fine-tuning, rises from approximately 9.1~minutes in 2D (where the MSCP split calibration requires 429.37~seconds) to about 1.3~hours for the 3D wing (with the MSCP phase taking 3,989.23~seconds). Crucially, once calibrated, the online prediction phase incurs a very low computational cost, requiring only 1.25~ms for a 2D snapshot and merely 8.57~ms for the 3D configuration. This high computational throughput confirms the framework's capability to deliver efficient, uncertainty-aware aerodynamic evaluations that seamlessly integrate into iterative optimization loops.

\subsection{Latent Space Analysis}
A central design constraint of this framework is the alignment of the latent space dimensionality with the number of governing physical parameters. This strategy relies on the premise that the intrinsic variability of the data can be effectively compressed into a representation commensurate with the system's true degrees of freedom \cite{frances2024toward}.

Beyond data compression, this topological isomorphism significantly enhances model interpretability. It allows for the identification of direct correlations between the abstract latent coordinates and the physical input parameters. Crucially, given the unsupervised nature of the autoencoder, these relationships are not explicitly imposed via labels but are autonomously discovered based solely on the structural patterns extracted from the pressure fields.

The following sections analyze the structure of the latent spaces generated by the proposed autoencoder for the two datasets under study.
\begin{figure}
    \centering
    \includegraphics[width = 8.2 cm]{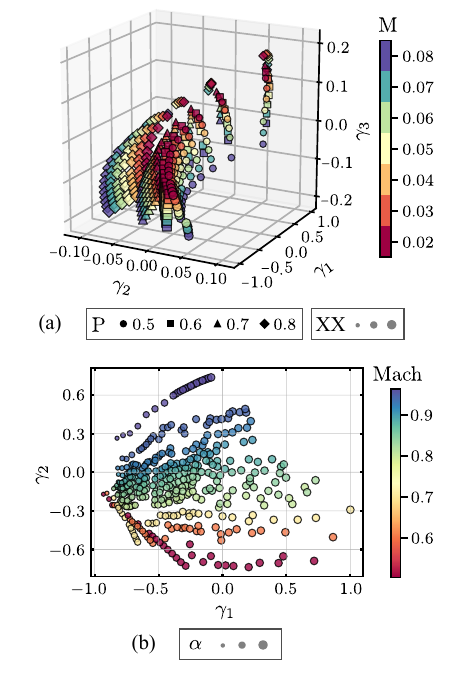}
    \caption{Latent space representations: (a) \textit{Airfoils} (40\% LF/HF) via camber M (color), position P (shape), thickness XX (size); (b) \textit{Wing} via Mach (color) and $\alpha$ (size).}
    \label{latent_spaces}
\end{figure}

\subsubsection{Airfoils Database}
The \textit{Airfoils database} comprises four-digit NACA airfoils fully parameterized by three independent geometric variables ($M,P,XX$). Consequently, it is postulated that a three-dimensional latent space ($d=3$) is optimal to encapsulate the intrinsic physical variability of the system.

\autoref{latent_spaces}(a) visualizes the learned manifold, revealing a highly structured decomposition where all latent coordinates exhibit an inverse dependency on their respective geometric parameters. The first latent dimension ($\gamma_1$) is driven by the maximum thickness ($\textit{XX}$), where the symbol size notably decreases along the axis. Similarly, the position of maximum camber ($P$) follows an inverse ordering along the second dimension ($\gamma_2$). Finally, the maximum camber ($M$) presents a negative correlation with the vertical dimension ($\gamma_3$), evidenced by the gradient transitioning from high-camber blue tones at the bottom to low-camber red tones at the top. This organized topology confirms that the model has successfully learned to disentangle and encode the fundamental relation between the pressure field and the geometric features of the airfoil without supervision.

\subsubsection{Wing Database}
The \textit{Wing database} simulations are driven by two governing flow parameters: the Mach number and $\alpha$. Consistent with the proposed methodology, the latent space dimensionality is therefore fixed at $d=2$.

\autoref{latent_spaces}(b) depicts the resulting latent space, which exhibits an unequivocal correspondence between the learned variables and the simulation parameters. The color gradient maps the Mach number, while the symbol size scales with the $\alpha$. The projection reveals a clear organization: $\alpha$ presents a distinct positive correlation with the first latent dimension ($\gamma_1$), evidenced by the progressive increase in marker size along the horizontal axis. Conversely, the Mach number exhibits a direct correlation with the second latent dimension ($\gamma_2$). This result demonstrates the model's ability to construct a meaningful, physics-aligned coordinate system from the data.

This highly structured organization, where the latent topology aligns with the physical parameters, is particularly advantageous for developing parametric surrogate frameworks, as explored in \cite{frances2024toward, nieto2024multifidelity, Castellanos2023isomap}. Such a structure facilitates the establishment of a direct regression mapping from the flight conditions ($Mach, \alpha$) to the latent coordinates, effectively obviating the need for new LF snapshots during the inference phase. While this approach would introduce an additional source of regression error, it offers a potential for significant computational savings by bypassing the generation of LF data entirely, enabling predictions driven solely by the simulation parameters.

\section{PERFORMANCE EVALUATION AND DISCUSSION}
This section presents the empirical validation of the proposed MF framework, evaluating both its predictive accuracy and the statistical data of the generated uncertainty quantification. The experimental campaign is structured around two representative case studies spanning different complexity levels: the \textit{Airfoils database} (2D), utilized to conduct a controlled sensitivity analysis regarding input resolution and architectural constraints, and the \textit{Wing database} (3D), which targets a representative industrial scenario characterized by complex transonic flows and severe data scarcity. In both instances, performance is evaluated through quantitative error metrics, conformal coverage statistics, and detailed qualitative inspections of the reconstructed flow physics.

To quantify predictive performance, three complementary error metrics are employed: the Mean Absolute Error (MAE), the Root Mean Square Error (RMSE), and the coefficient of determination ($R^2$). MAE provides an average measure of the pointwise deviation between prediction and ground truth, offering a direct interpretation of the typical regression error. RMSE penalizes larger discrepancies more heavily, making it particularly sensitive to localized errors that may arise near discontinuities or strong gradients. In contrast, $R^2$ evaluates the global ability of the model to reproduce the overall variance of the HF data. Together, these metrics capture different aspects of accuracy, thereby enabling a comprehensive assessment of the proposed framework.

In parallel, the reliability and efficiency of the uncertainty quantification provided by the MSCP framework are assessed through three statistical metrics. First, Nominal Coverage is computed as the fraction of test cases where the entire output vector is simultaneously contained within the predicted uncertainty margins, serving as a strict criterion for multivariate validity. Although the proposed framework ultimately emphasizes pointwise utility for engineering decision-making, reporting nominal coverage remains important as an empirical diagnostic of how far the aggregated MSCP bands depart from the strict simultaneous coverage behavior associated with standard split conformal prediction. This is complemented by Pointwise Coverage, which calculates the aggregate percentage of individual components of the $C_p$ vector falling within the bounds, offering a more granular assessment of local reliability. To evaluate the practical usefulness of the uncertainty model, the average band width is analyzed, defined as the average physical distance between the upper and lower uncertainty bounds of the predicted output. Additionally, the standard deviation of this width is reported to quantify its spatial variability, reflecting the extent to which the uncertainty margin deviates from the average to accommodate local flow fluctuations.

\subsection{Airfoils Database}
The evaluation for this database is performed across varying LF input resolutions to assess the efficacy and robustness of the proposed MFAE. The quantitative analysis includes deterministic prediction metrics, conformal prediction coverage, and a qualitative inspection of the reconstructed fields.
\begin{table*}
    \caption{\label{tab:model_metrics} Model performance ($MAE$, $RMSE$, $R^2$) on the \textit{Airfoils database} for various LF/HF ratios, evaluated using test and complementary test datasets.}
    \centering
    \renewcommand{\arraystretch}{1}
    \begin{tabular}{c ccc c ccc}
        \toprule
        & \multicolumn{3}{c}{Test} & & \multicolumn{3}{c}{Complementary Test} \\
        \cmidrule(lr){2-4} \cmidrule(lr){6-8}
        LF/HF Dim (\%) & MAE & RMSE & $R^{2}$ & & MAE & RMSE & $R^{2}$ \\
        \midrule
        15  & 0.034 & 0.110 & 0.995 & & 0.039 & 0.100 & 0.996 \\
        25  & 0.027 & 0.080 & 0.997 & & 0.032 & 0.079 & 0.997 \\
        40  & 0.025 & 0.070 & 0.998 & & 0.028 & 0.070 & 0.998 \\
        60  & 0.018 & 0.055 & 0.999 & & 0.020 & 0.054 & 0.999 \\
        85  & 0.019 & 0.064 & 0.998 & & 0.021 & 0.063 & 0.998 \\
        100 & 0.025 & 0.090 & 0.996 & & 0.022 & 0.064 & 0.998 \\
        100 (*) & 0.018 & 0.053 & 0.999 & & 0.021 & 0.058 & 0.999 \\
        \bottomrule
    \end{tabular}
\end{table*}

\autoref{tab:model_metrics} summarizes the regression metrics. It is noteworthy that this specific assessment isolates the deterministic predictive capability of the autoencoder, independent of the uncertainty quantification provided by the MSCP.

A global analysis of the metrics reveals a consistently robust performance across the tested configurations. The coefficient of determination ($R^2$) remains close to unity in all scenarios, corroborating the model’s strong predictive capability and its ability to predict the trends of the HF data. This result suggests that the model not only captures the dominant sources of variability but also provides an accurate representation of the overall system behavior, even when driven by coarsened inputs.

Correlating input dimensionality with predictive precision reveals several key trends. As expected, performance error is most pronounced at the lowest resolution (LF/HF 15\%). This behavior is consistent with the premise that an excessively reduced input dimensionality fails to provide the encoder, during the pre-training phase, with sufficient information about the local flow features. This lack of information in the latent representation limits the decoder's ability to accurately reconstruct the complex structure of the HF data in the final training phase.

However, the results also indicate a clear saturation point. Beyond an input dimensionality of 60\%, marginal gains in predictive accuracy diminish significantly. This suggests that at 60\% resolution, the encoder has already extracted the main physical features contained in the LF data; further increasing nodal density adds no new transferable information to the learning process.

Interestingly, the configuration in which the input dimensionality is identical to the output produces a notable increase in error, reversing the trend of improvement. This seemingly counterintuitive result can be explained by our model's architecture and the nature of transfer learning. In the standard 100\% setup, the adaptation layer (typically used for up-scaling) is omitted since dimensions match. This forces the decoder, pre-trained with LF data, to map the latent space directly to the HF output. The pre-trained decoder has deeply learned the inherent biases of the LF inviscid data. The absence of additional layers, trained from scratch in the final training phase, reduces the model's plasticity and, consequently, the model struggles to adjust to the viscous effects present in the HF data.

To validate this hypothesis, a control experiment (denoted as $100\% (*)$) was conducted. Here, an additional hidden layer was forcibly inserted into the architecture and trained from scratch during the transfer learning phase, even though it was not required for dimensional compatibility. The results confirm that the inclusion of this additional layer led to a notable improvement in performance metrics, yielding an RMSE of 0.053. It can therefore be concluded that the introduction of post-decoder layers fulfills a dual function: beyond handling dimensional up-scaling, these layers operate as corrective modules that provide the model with the necessary flexibility to compensate for biases inherited from the LF pre-training, thereby enabling a more effective adaptation to the HF data.
\begin{table*}[]
    \caption{\label{coverage_mscp} MSCP nominal and pointwise coverage and calibrated band width (mean $\pm$ std) for \textit{Airfoils database} LF/HF ratios, using RMSR and $L_\infty$ scoring functions.}
    \renewcommand{\arraystretch}{1.2}
    \resizebox{\textwidth}{!}{
    \begin{tabular}{ccccccccc}
        \hline
        \multicolumn{1}{l}{} &  & \multicolumn{3}{c}{RMSR Coverage} &  & \multicolumn{3}{c}{$L_\infty$ norm Coverage} \\ \cline{3-5} \cline{7-9} 
        \multicolumn{1}{l}{} & LF/HF Dim (\%) & Nominal & Pointwise & Band width + Std &  & Nominal & Pointwise & Band width + Std \\ \hline
        \multirow{7}{*}{\rotatebox[origin=c]{90}{Test}} & 15  & 0.75 & 0.96 & [0.17, 0.14] & & 0.95 & 0.99 & [0.50, 0.39] \\
         & 25  & 0.60 & 0.96 & [0.16, 0.14] & & 0.95 & 0.99 & [0.50, 0.43] \\
         & 40  & 0.80 & 0.97 & [0.15, 0.11] & & 0.95 & 0.99 & [0.50, 0.36] \\
         & 60  & 0.85 & 0.97 & [0.14, 0.10] & & 0.95 & 0.99 & [0.45, 0.32] \\
         & 85  & 0.90 & 0.98 & [0.14, 0.10] & & 0.95 & 0.99 & [0.46, 0.30] \\
         & 100 & 0.85 & 0.97 & [0.18, 0.22] & & 0.95 & 0.99 & [0.54, 0.57] \\
         & 100 (*) & 0.75 & 0.97 & [0.12, 0.09] & & 0.95 & 0.99 & [0.25, 0.18] \\ \hline \hline
        \multirow{7}{*}{\rotatebox[origin=c]{90}{\begin{tabular}[c]{@{}c@{}}Complementary \\ Test\end{tabular}}} & 15 & 0.66 & 0.93 & [0.17, 0.14] &  & 0.91 & 0.98 & [0.50, 0.39] \\
         & 25 & 0.59 & 0.93 & [0.16, 0.14] &  & 0.93 & 0.98 & [0.50, 0.43] \\
         & 40 & 0.78 & 0.95 & [0.15, 0.11] &  & 0.93 & 0.99 & [0.50, 0.36] \\
         & 60 & 0.80 & 0.96 & [0.14, 0.10] &  & 0.92 & 0.99 & [0.45, 0.32] \\
         & 85 & 0.84 & 0.96 & [0.14, 0.10] &  & 0.93 & 0.99 & [0.46, 0.30] \\
         & 100 & 0.78 & 0.98 & [0.18, 0.22] &  & 0.97 & 0.99 & [0.54, 0.57] \\
         & 100 (*) & 0.69 & 0.95 & [0.12, 0.09] &  & 0.91 & 0.98 & [0.25, 0.18] \\ \hline
    \end{tabular}}
\end{table*}

The coverage metrics derived from the MSCP method are detailed in \autoref{coverage_mscp}. In all scenarios, the models were calibrated within each individual data split to target a nominal significance level of $\delta=0.1$, implying a target simultaneous coverage of 90\%. Regarding nominal coverage, a marked divergence emerges between the non-conformity functions. The use of the $L_\infty$ norm consistently meets and even notably exceeds the desired target coverage. In contrast, the metric based on the RMSR consistently results in simultaneous under-coverage. This discrepancy is driven by the nature of the scoring functions. The $L_{\infty}$ norm calibrates the interval based on the worst-case individual error, naturally yielding conservative radii large enough to bound all dimensions. Conversely, the RMSR averages errors, producing significantly tighter intervals. Crucially, since our MSCP framework aggregates results using the median to prioritize stability, strict nominal coverage is not theoretically guaranteed. Consequently, the high coverage observed for $L_{\infty}$ is an empirical result: its inherent conservatism generates bands broad enough to withstand the median aggregation, while the narrower RMSR intervals become insufficient to capture the full output vector.

However, the high nominal coverage of the $L_\infty$ norm comes at the cost of a severe practical limitation. As can be observed, the uncertainty bands generated by this score are, on average, twice as wide as those generated by the RMSR. From an engineering design perspective, this situation is suboptimal; an uncertainty margin that is excessively large, while statistically valid, diminishes the practical utility of the surrogate, as the range of potential solutions becomes too vast to inform decision-making. Consequently, priority is given to pointwise coverage as a more pragmatic performance indicator.

From this perspective, the analysis of \autoref{coverage_mscp} yields a critical insight: the pointwise coverage rates between the two metric types are remarkably close. Taking the 60\% LF/HF dimensionality configuration as a reference, the pointwise coverage obtained with RMSR decreases by only 2\% compared to the $L_\infty$ score. This marginal reduction in individual point reliability is offset by substantial gains in informativeness, as the mean band width and its standard deviation are reduced by approximately 70\%. Therefore, for engineering design purposes, pointwise coverage emerges as a better performance indicator. It enables the selection of an uncertainty quantification method that balances efficiency and reliability, generating uncertainty bands that are narrow enough to be actionable while ensuring that the vast majority of critical flow features are reliably bounded.

\begin{figure}[h!]
    \centering
    \includegraphics[width = 7.2 cm]{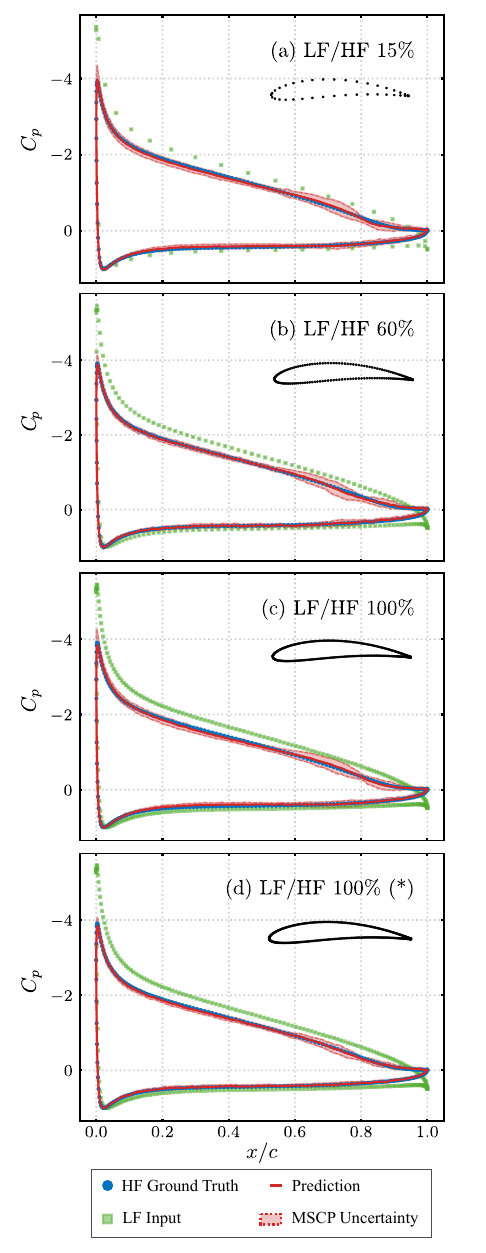}
    \caption{Predictive $C_p$ for NACA 6513 (\textit{Airfoils database}) under varying LF/HF mesh density ratios: (a) 15\%, (b) 60\%, (c) 100\%, and (d) 100\% (*).}
    \label{quadrio_cp}
\end{figure}

To complement the quantitative analysis, \autoref{quadrio_cp} presents a visual comparison of the prediction results and uncertainty bands for a specific test case (NACA 6513 airfoil) under three distinct LF/HF dimensionality configurations (15\%, 60\%, and 100\%). At the lowest resolution (15\% LF/HF, \autoref{quadrio_cp}(a)), the limitations imposed by sparse input resolution become visible. Although the prediction captures the global trend of the pressure field, it exhibits a certain bias relative to the HF data. The CP framework responds appropriately to this increased uncertainty. The confidence bands are slightly wider across the entire chord compared to cases with higher input resolutions, with a more pronounced widening still observed in the suction peak region ($x/c \approx 0.1$). Even though the peak prediction itself remains reasonably accurate, the band's width is not based solely on this single localized error. Instead, it reflects the aggregation of residuals across the $B = 30$ calibration splits of the MSCP, correctly indicating that the model recognizes this zone as inherently more difficult to predict under strict data constraints.

In contrast, \autoref{quadrio_cp}(b) (60\% LF/HF) depicts the configuration where the architecture reaches its performance saturation point. The performance improvement is evident. The model's prediction aligns with high precision with the HF data along both surfaces of the airfoil. This predictive accuracy translates directly into high model confidence, resulting in uncertainty bands that are narrower compared to the 15\% case. This finding validates the metrics reported in \autoref{coverage_mscp}, demonstrating that the model achieves this tightness while maintaining robust pointwise coverage.

Finally, the analysis of the full-resolution cases reveals a critical architectural insight regarding model plasticity. As shown in \autoref{quadrio_cp}(c) (100\% LF/HF), matching input and output dimensions leads to a counter-intuitive increase in prediction error. This phenomenon stems from the omission of the adaptation layer, which forces the fine-tuning process to rely exclusively on the pre-trained decoder. Lacking the additional learnable parameters of an up-scaler, the model fails to capture the viscous effects intrinsic to the HF data. However, this limitation is effectively resolved in the corrected experiment shown in \autoref{quadrio_cp}(d) ($100\% (*)$), where an additional trainable hidden layer was enforced. This modification eliminates the systematic bias, restoring agreement with the HF data points. Crucially, a comparison between this configuration and the 60\% case (\autoref{quadrio_cp}(b)) reveals that the regression predictions are very similar, with only a reduction in the uncertainty band width within the regions of highest variability. This reinforces the information saturation hypothesis derived from the quantitative analysis: once the LF input resolution is sufficient to capture the dominant flow physics, further increasing input density yields only marginal benefits to reconstruction fidelity.

A common characteristic is identified across all three configurations: a consistent widening of the confidence bands in the rear region of the airfoil (approximately $x/c > 0.6$). A detailed examination of the data reveals a physical discrepancy in this zone; the RANS data exhibits a curvature change that is not present in the LF data. Because the predictive model is strongly influenced by the LF data during pre-training, it slightly under-resolves this local curvature change, resulting in a noticeable artificial undulation in the prediction. However, this observation highlights the value and robustness of the uncertainty quantification framework. The CP method correctly detects this zone of systematic discrepancy, accumulated during the $B$ calibrations, and locally widens the uncertainty to maintain coverage. This not only validates the safety of the method but also signals to the engineer a specific region where the inter-fidelity correlation is weak, providing a warning that the point prediction is less reliable.

\subsection{Wing Database}
In this section, the predictive performance of the proposed framework is evaluated using the \textit{Wing database}, described in \ref{Garteur_database}. In contrast to the parametric analysis performed with the \textit{Airfoils database}, this study focuses on a fixed configuration that simulates a highly representative aeronautical scenario.

The computational challenge presented here is twofold and severe. First, the model is tasked with predicting a high-dimensional field ($D_{HF} = 49{,}574$ nodes) while relying on an extremely sparse HF training budget ($N=65$ snapshots). Second, the LF input, despite being more abundant, is provided at a coarse spatial resolution ($D_{LF} = 4{,}000$). Consequently, the primary objective is to quantify the framework's effectiveness and robustness when applied to a complex three-dimensional aerodynamic problem subject to the stringent data scarcity constraints typical of advanced engineering design.

\begin{table}
    \caption{\label{tab:model_metrics_garteur} Model performance ($MAE$, $RMSE$, $R^2$) on the \textit{Wing database} for various LF/HF ratios, evaluated using test and complementary test datasets.}
    \centering
    \begin{tabular}{ccc c ccc}
        \hline
        \multicolumn{3}{c}{Test} & & \multicolumn{3}{c}{Complementary test}\\
        \cline{1-3} \cline{5-7}
        MAE & RMSE & $R^{2}$ & & MAE & RMSE & $R^{2}$\\
        \hline
        0.046 & 0.084 & 0.965 & & 0.049 & 0.087 & 0.963\\
        \hline
    \end{tabular}
\end{table}

\autoref{tab:model_metrics_garteur} summarizes the model's global performance metrics, evaluated on both the standard Test set and the Complementary Test set. An excellent concordance is observed between the two sets, with nearly identical values for MAE, RMSE, and an $R^2$ coefficient of 0.965 versus 0.963. This similarity is fundamental, as it confirms that the Test set constitutes a statistically representative sample of the global generalization performance, despite its limited size. Such validation is critical in an industrial setting, where typically only a very small subset of HF data is available for testing, ensuring that metrics derived from a few samples serve as reliable indicators of global performance.

Based on the Test set results, the model demonstrates a robust predictive capability. The model achieves a high determination coefficient ($R^2 = 0.965$), while the MAE and RMSE, at 0.046 and 0.084, respectively, are considered low for the analyzed $C_p$ range. The significance of these findings is amplified when contextualized with existing literature. Prior studies \cite{frances2024toward, castellanosassessment, hines2022data}, that have utilized this same \textit{Wing database} for single-fidelity regression tasks, trained solely on HF data, report error metrics of a comparable magnitude. The fact that the proposed MF model, trained with a drastically reduced HF set of only 65 snapshots, achieves global accuracy similar to that of single-fidelity models requiring much denser training underscores the extraordinary data efficiency of the strategy. This validates the approach as a viable solution for reducing the computational burden of aerodynamic characterization without compromising global accuracy.

\begin{table*}
    \caption{\label{tab:coverage_mscp_combined} MSCP nominal and pointwise coverage and calibrated band width (mean $\pm$ std) for \textit{Wing database}, using RMSR and $L_\infty$ scoring functions.}
    \centering
    \begin{tabular}{cccc c ccc}
    \hline
     & \multicolumn{3}{c}{RMSR Coverage} & & \multicolumn{3}{c}{$L_\infty$ norm Coverage} \\
    \cline{2-4} \cline{6-8}
    & Nominal & Pointwise & Band Width + Std & & Nominal & Pointwise & Band Width + Std \\
    \hline
    Test & 0.05 & 0.95 & [0.34, 0.17] & & 0.41 & 0.99 & [0.90, 0.46] \\
    Complementary Test & 0.003 & 0.95 & [0.34, 0.17] & & 0.33 & 0.99 & [0.90, 0.46] \\
    \hline
    \end{tabular}
\end{table*}

Table~\ref{tab:coverage_mscp_combined} presents the coverage results for the MSCP method, where each individual data split is calibrated for a target nominal coverage of 90\% ($\delta = 0.1$). Consistent with the Airfoils analysis, the evaluation distinguishes between non-conformity functions and coverage definitions.

Regarding the RMSR, the simultaneous nominal coverage is negligible (5\%). Consistent with the observations in the \textit{Airfoils database}, this confirms that a radius derived from an averaged metric is structurally insufficient to envelop the complete $C_p$ vector. This limitation is intensified by the scale of the problem, considering that the wing $C_p$ vector comprises 49,574 components compared to only 260 in the airfoil cases. Under such high dimensionality, the resulting band is too narrow to prevent at least one point from falling outside the margins in almost all test cases. This behavior is consistent with the underlying formulation of the RMSR-based conformal set: its exact guarantee is naturally associated with an ellipsoidal region in the full output space of the reconstructed $C_p$ field, whereas representing it as a hyperrectangular band for visualization does not preserve simultaneous vector-level coverage. Even the $L_{\infty}$ norm, which is the metric most directly aligned with simultaneous hyperrectangular coverage, fails to meet the target, reaching only 41\% in terms of nominal coverage. This outcome clearly illustrates the specific trade-off imposed by the MSCP aggregation in complex flow regimes. The \textit{Wing dataset} contains strong nonlinearities, such as shock waves, which induce local deviations in the predictions. While these discontinuities generate large calibration radii in specific data splits, the median aggregation attenuates their impact to preserve stability, thereby prioritizing robust pointwise utility. Consequently, unlike in the 2D case, the resulting stable band is too narrow to encapsulate the most severe local gradients.

Evaluating pointwise coverage reveals a clearer perspective. The $L_\infty$ norm yields an excessively conservative pointwise coverage of 99\%. This high coverage is a direct consequence of an overly wide uncertainty band; its mean width (0.90) is 2.6 times larger than that of the RMSR (0.34). In a practical engineering design setting, such broad margins provide no actionable value for decision-making. In contrast, the RMSR achieves a pointwise coverage of 95\%, providing a highly satisfactory level of reliability while ensuring much narrower, practically useful prediction bands. Therefore, the RMSR offers a far superior trade-off between statistical protection and practical utility.

Finally, comparing the Test and Complementary Test sets reveals that band width and pointwise coverage remain perfectly stable. The only discrepancy appears in the nominal coverage for RMSR (5\% vs 0.3\%). This is identified as a statistical artifact of the sample size. In the test set ($N=22$), a single covered case represents $\approx 5\%$ of the sample. The Complementary set ($N=370$) provides a finer resolution, confirming that true simultaneous coverage for the RMSR score in this high-dimensional space is, as theoretically predicted, asymptotically zero.

\begin{figure}[ht]
    \centering
    \includegraphics{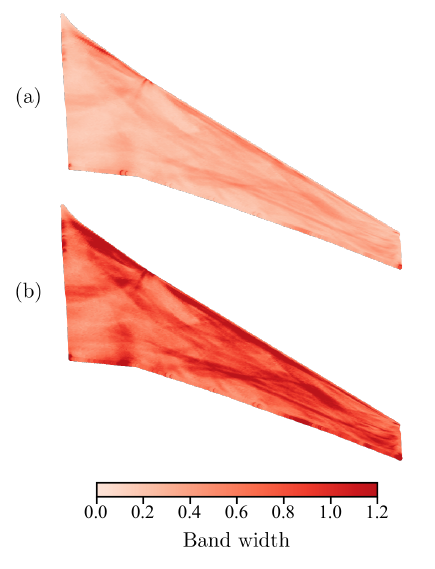}
    \caption{Mean uncertainty band width obtained using (a) the RMSR and (b) the $L_\infty$ norm scoring functions.}
    \label{conformal_bandwidth_garteur}
\end{figure}

\autoref{conformal_bandwidth_garteur} illustrates the spatial distribution of the calibrated prediction band widths. This visual assessment corroborates the quantitative evidence presented in \autoref{tab:coverage_mscp_combined}: while the topological structure of the uncertainty fields remains qualitatively consistent between methods, the intervals generated by the $L_\infty$ norm exhibit a substantially greater magnitude across the entire wing surface.

A critical insight provided by these contours is the physical correlation of the uncertainty margins. Both methodologies successfully localize distinct regions of high uncertainty that align with zones of reduced predictive fidelity. These areas correspond to flow regimes dominated by strong adverse pressure gradients, specifically the suction peak recovery near the leading edge and, most prominently, the shock waves. Given that the underlying regression model exhibits higher residuals when capturing these sharp discontinuities, the coherent inflation of the uncertainty bands in these physically critical locations validates the framework's ability to spatially map the underlying error structure.

\begin{figure}[ht]
    \centering
    \includegraphics[width = 8.2 cm]{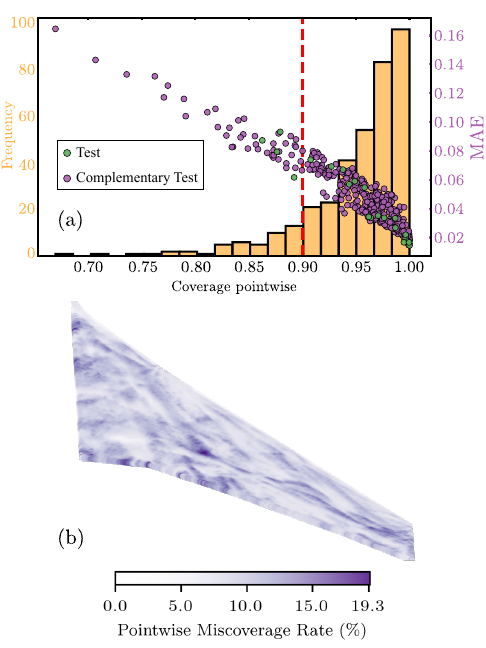}
    \caption{(a) MSCP performance for \textit{Wing database}. The scatter plot shows $MAE$ vs. Pointwise Coverage. The orange histogram displays coverage distribution.  (b) Individual predictions outside uncertainty bands.}
    \label{coverage_error}
\end{figure}

\autoref{coverage_error}(a) provides a granular analysis of the MSCP framework's performance, examining the distribution of error and coverage across the entire evaluation corpus. The frequency analysis reveals that the vast majority of snapshots surpass 90\% of the pointwise coverage. This reaffirms that the excellent global pointwise coverage reported in \autoref{tab:coverage_mscp_combined} is a robust statistical indicator and not merely an artifact of averaging high and low performing cases. Furthermore, the scatter plot reveals a distinct inverse correlation: as a snapshot's MAE increases, its pointwise coverage decreases significantly. This behavior is a direct consequence of the employed methodology, which generates a calibrated uncertainty band that varies spatially yet remains independent of the model features. While this band is sufficient for the majority of cases, it proves inherently inadequate for covering prediction outliers. These high-error and low-coverage cases correspond to the most physically complex scenarios, such as high Mach number and $\alpha$ conditions featuring intense shock waves, or to cases located at the edges of the training data's convex hull, where the model is forced to extrapolate. Overall, this figure not only validates the robustness of the coverage for most of the domain but also serves as a diagnostic tool. It precisely identifies the conditions where the base regression model fails and, consequently, where the uncertainty band underestimates the true error.

To assess the localized reliability of the uncertainty quantification, \autoref{coverage_error}(b) illustrates the spatial distribution of the pointwise miscoverage rate across the wing surface, computed over the combined test and complementary test datasets. This metric quantifies the percentage of test instances where the HF reference value at any given coordinate falls outside the predicted uncertainty bands. The highest miscoverage rates, reaching a localized maximum of 19.3\%, are primarily concentrated within the regions swept by the shifting shock wave under varying flight conditions, though some isolated errors are also visible near the trailing edge. Nevertheless, the magnitude of these percentages indicates that the reference data exceeds the uncertainty margins only under specific operating conditions characterized by strong shock waves, where regression errors are inherently elevated. Globally, the results demonstrate that the predicted uncertainty margins are well-calibrated, as evidenced by the low miscoverage rates maintained across the vast majority of the wing surface.

\begin{figure*}[!t]
    \centering
    \includegraphics[width = 0.90\textwidth]{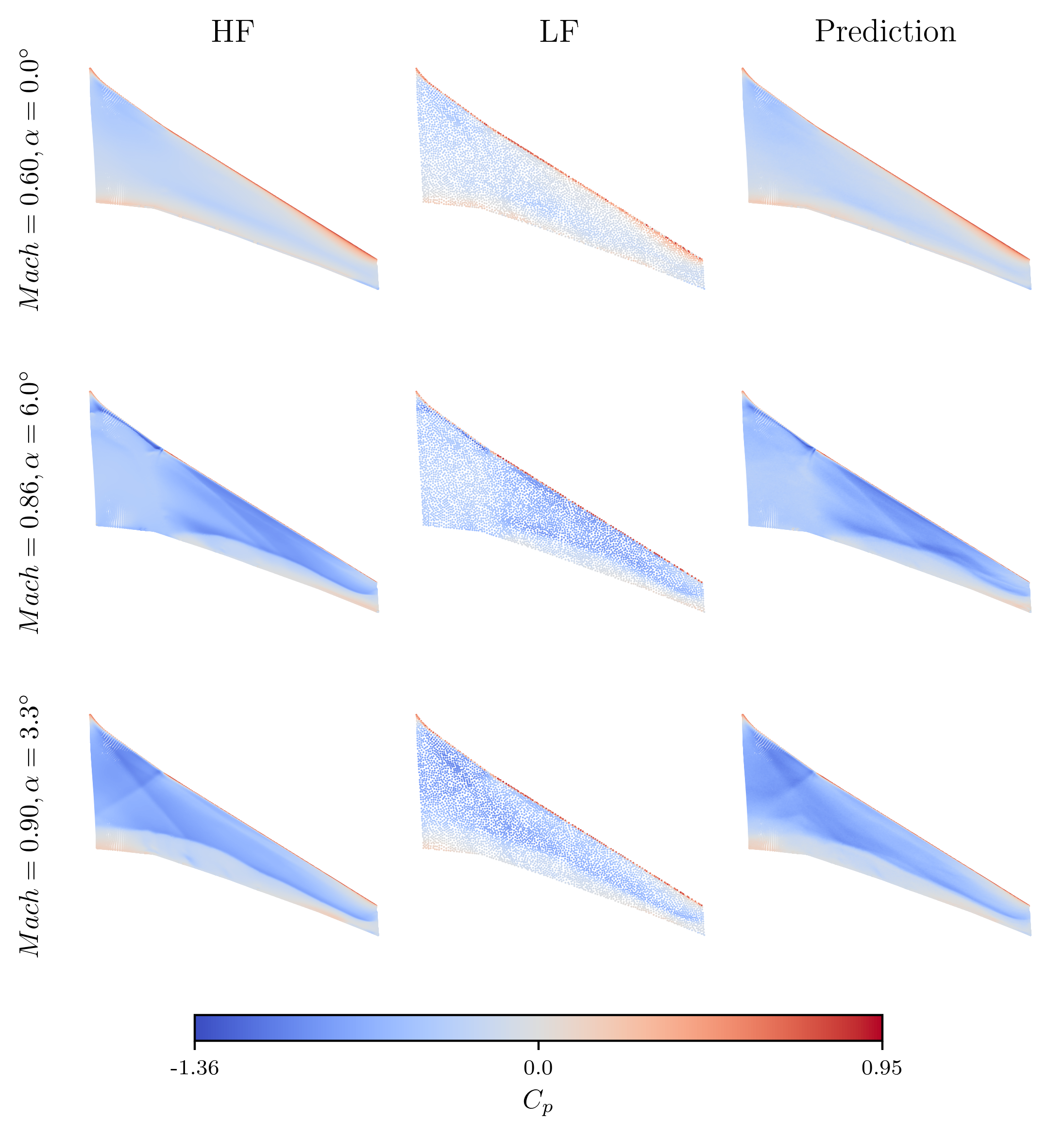}
    \caption{MF reconstruction: reference $C_p^{HF}$ (left), input $C_p^{LF}$ (middle), and predicted $\tilde{C}_p$ (right) for test cases ranging from attached flow to shock-dominated regimes.}
    \label{wing_prediction}
\end{figure*}

\begin{figure*}[!t]
    \centering
    \includegraphics[width = 0.90\textwidth]{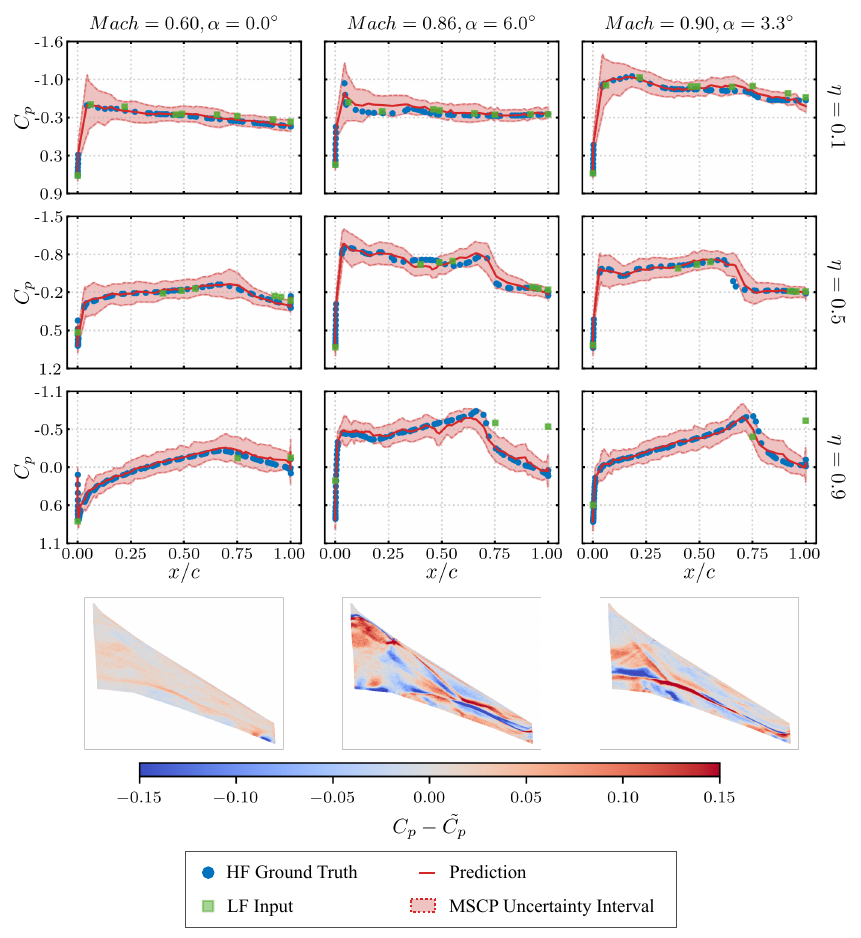}
    \caption{Local performance: wing surface prediction error and chordwise $C_p$ at $\eta \in \{0.1, 0.5, 0.9\}$ for three representative flight conditions.}
    \label{garteur_sections}
\end{figure*}

To complement the global metrics, a detailed qualitative analysis is conducted on three flight conditions representative of the design space. \autoref{wing_prediction} illustrates the qualitative comparison of the $C_p$ distribution over the wing surface (Ground Truth HF, Input LF, and Predicted HF), while \autoref{garteur_sections} provides an in-depth local analysis, displaying chordwise sectional cuts (at $\eta = y/(b/2)=\{0.1, 0.5, 0.9\}$, being $b/2$ the semi-wing span), the generated uncertainty bands, and absolute error maps ($C_p - \tilde{C}_p$) projected onto the geometry.

In the first condition studied, corresponding to a subsonic regime at a low angle of attack ($Mach = 0.6, \alpha = 0.0^\circ$), the flow remains largely attached and devoid of strong discontinuities. An excellent agreement is observed between the prediction and the HF reference. The error map in \autoref{garteur_sections} reveals a highly homogeneous residual distribution, close to zero across the vast majority of the surface. The most notable discrepancies are localized at the trailing edge near the wing tip and in the final third of the chord,  coinciding with the pressure recovery region. Examination of the cross-sections confirms that the prediction faithfully reproduces the pressure profile. Remarkably, given the flow stability, the predicted uncertainty bands successfully encapsulate all observed data points, validating the model's calibration in linear regimes.

The second condition analyzed represents a significantly more demanding aerodynamic scenario ($Mach = 0.86, \alpha = 6.0^\circ$), characterized by the presence of a well-defined shock wave on the upper surface. Although the prediction correctly captures the general flow topology and the shock wave position, slight numerical oscillations are visible in the reconstructed surface. The error map clearly identifies the shock wave location as the main source of epistemic uncertainty, along with regions of adverse pressure gradients near the wing root. However, the sectional analysis reveals one of the proposed methodology's greatest strengths: its bias correction capability. As observed in section $\eta = 0.9$, the LF input data exhibit a massive structural discrepancy with respect to the HF ground truth, erroneously predicting the flow physics. Nevertheless, the model successfully decouples itself from this bias learned during the pre-training phase and, through fine-tuning, recovers the correct HF distribution. Furthermore, the uncertainty bands demonstrate proper performance for this regime, maintaining robust empirical coverage.

Finally, a critical condition is analyzed ($Mach = 0.9, \alpha = 3.3^\circ$), exhibiting complex nonlinear phenomena, including an intense shock wave and wing-fuselage junction interactions. The prediction succeeds in defining the structure of the main shock wave with remarkable sharpness. However, as evidenced by the error map, the largest discrepancies are concentrated in the wing root region, where shock interactions generate highly complex fluid dynamics that are difficult to capture with the proposed model and a limited training set. Detailed sectional analysis (specifically at $\eta = 0.5$) reveals an intrinsic limitation of interval prediction in the presence of extreme gradients. At the exact location of the shock wave, the actual $C_p$ value falls outside the uncertainty bands. This occurs because a minimal positional error in the shock wave prediction translates into a very high vertical $C_p$ error; although the MSCP method widens the band in areas of higher model error, the gradient is so abrupt that the interval fails to cover the point-wise ground truth value. Notwithstanding this local singularity, the global coverage for the case remains satisfactory, and the model proves competent in predicting nonlinear physics across the span.

\section{CONCLUSIONS}
This work has presented a comprehensive deep learning framework for MF aerodynamic data fusion, designed to overcome the constraints of HF data scarcity in aerospace engineering. By integrating an autoencoder-based transfer learning architecture with an MSCP strategy, the proposed methodology enables the robust prediction of complex aerodynamic fields while providing reliable uncertainty quantification.

The results obtained from both the parametric airfoil dataset and the transonic wing configuration demonstrate the efficacy of the transfer learning strategy. The latent space analysis confirmed that the encoder, pre-trained on abundant LF data, successfully learns a structured and interpretable representation of the flow physics, autonomously correlating latent variables with geometric and flight parameters. A critical methodological finding is the necessity of a trainable adaptation layer during the fine-tuning phase. The analysis revealed that even when input and output dimensionalities match, the inclusion of this intermediate layer is mandatory to decouple the model from the inductive biases inherent to LF data. This architectural modification proved essential, effectively eliminating systematic errors and reducing the reconstruction error.

Regarding the efficiency of the framework, the results on the GARTEUR XRF1 wing highlight a substantial reduction in computational cost. The model achieved $R^2=0.965$ and $MAE=0.046$. The study employed a total HF budget of 87 cases (65 training, 22 testing). When compared to the complete database of 435 flight conditions, this represents a utilization of only 20\% of the potential data. Consequently, the proposed framework enables the reconstruction of the global $C_p$ field with a predictive precision comparable to single-fidelity models while drastically reducing the HF computational burden, thereby offering a scalable pathway to accelerate design cycles.

The proposed MSCP framework performs robustly in the present data-limited settings, effectively mitigating the sensitivity to random data partitions that typically affect single-split validation procedures. To ensure the practical utility of the surrogate, the methodology prioritizes the generation of tight pointwise uncertainty bands with strong empirical coverage. This decision is driven by the necessity of providing actionable margins that are sufficiently narrow to inform engineering decisions. By adopting the RMSR combined with median aggregation, the framework balances statistical rigor with utility, reducing the average band width by approximately 50\% relative to the $L_\infty$ norm. Crucially, this approach shows strong empirical calibration, with the true $C_p$ value falling within the predicted uncertainty bands for over 95\% of the wing surface grid points. Furthermore, these bands serve as a valuable diagnostic tool, automatically widening in regions of complex nonlinearities, such as shock waves, thereby flagging areas of reduced predictive confidence. Ultimately, the evidence provided in this work should be interpreted as an empirical validation of the MSCP strategy on two representative aerodynamic regression tasks of markedly different dimensionality, rather than as a universal validation of the method across arbitrary domains.

Beyond these results, the present framework establishes a novel pathway for deploying sophisticated deep learning models in multi-fidelity aerodynamic analysis while preserving robust uncertainty awareness throughout the design process. By enabling HF-accurate predictions from minimal HF data and providing actionable confidence bounds, the methodology strengthens the viability of MF surrogates for industrial decision-making and large-scale aerodynamic exploration. The framework is, in principle, applicable to other aerodynamic configurations and physical fields, provided LF–HF correlations exist. Nevertheless, the present study validates the methodology only for surface-pressure reconstruction, since $C_p$ is the common paired LF/HF observable consistently available across both benchmark datasets and because it provides a stringent target in the presence of shock-related and other strongly nonlinear aerodynamic features. Looking forward, substantial opportunities remain for further development. In particular, extending the approach toward adaptive or online transfer learning may allow the latent space to evolve beyond its LF-derived structure, capturing HF features that are not represented in the pre-training data. More broadly, the correlation and integration of latent spaces across multiple fidelities, or even across different disciplines, represent promising directions for building unified, physics-informed representations. Assessing the same framework on additional observables, such as wall-shear-related quantities or integral aerodynamic coefficients, remains a natural next step once suitably paired multi-fidelity datasets are available. These avenues point to a rich landscape for future research in scalable, uncertainty-aware aerodynamic modeling.

\section*{Acknowledgments}
This work has been supported by the TIFON project, ref. PLEC2023-010251/MCIN/AEI/ 10.13039/501100011033, funded by the Spanish State Research Agency. The authors would like to thank AIRBUS for providing the XRF1 database.

\bibliography{bibliography}

\newpage

\appendices

\section{Low-Fidelity Data Generation Methodologies} \label{Appendix_A}

This appendix details the techniques developed to generate LF datasets from a reference HF dataset. The development and validation of MF architectures, such as the autoencoder proposed in this work, require access to paired (LF-HF) datasets describing the same system. Frequently, in real application scenarios, HF data may be available, but a corresponding or simultaneously acquired LF dataset may not exist. To overcome this limitation and to systematically validate the correct functioning of the proposed methodology, it was necessary to implement a series of synthetic degradation procedures that allow for the synthetic and controlled generation of said LF data.

For this study, fidelity reduction has been interpreted and applied following two primary strategies, which can be employed independently or in combination. The first strategy is the reduction of data \textit{precision}. This approach seeks to decrease the numerical accuracy of the HF values without altering the dimensionality or structure of the data. This simulates the effect of using less precise measurement instruments, simulation models with lower arithmetic convergence, or simply data storage with lower numerical resolution. The result is a dataset that possesses the same shape and size as the HF original, but with degraded information quality.

The second strategy is the reduction of resolution, also known as \textit{downsampling}. This method does alter the data structure, as it reduces the dimensionality of the input vector. This is achieved by decreasing the information density, for example, by selecting a subset of points on a simulation mesh, reducing the sampling frequency in a temporal signal, or grouping features. This process directly mimics the result of a computational simulation performed with a coarser mesh or a simplified physical model, which intrinsically produces a less granular description of the system.

The following sections of this appendix will describe in detail the specific algorithms implemented to realize these two approaches, thereby providing the databases necessary for the training and validation of the MFAE.

\subsection{Reduction via Modal Filtering and Farthest Point Sampling (FPS)}
The first solution adopted, and the one used as the basis in the main body of this work, is a sequential combination of two techniques: POD to decrease data precision, followed by FPS to reduce the dimensionality or spatial resolution.

\subsubsection{Precision Reduction via Modal Filtering}
POD is a statistical Model Order Reduction framework that isolates an optimal orthogonal basis maximizing the variance of a snapshot matrix \cite{brunton2022data}. To implement this methodology, the HF data is organized into a snapshot matrix $\mathbf{X} = [\mathbf{c_p}{_1}, \mathbf{c_p}{_2}, \dots, \mathbf{c_p}{_N}] \in \mathbb{R}^{D \times N}$, where $D$ denotes the spatial dimension and $N$ the number of samples. Each column corresponds to a snapshot, defined as a column-wise vector of pressure coefficients $\mathbf{c_p}{_i}$ for the $i$-th snapshot.

\begin{figure}[H]
    \centering
    \includegraphics[width = 8.2 cm]{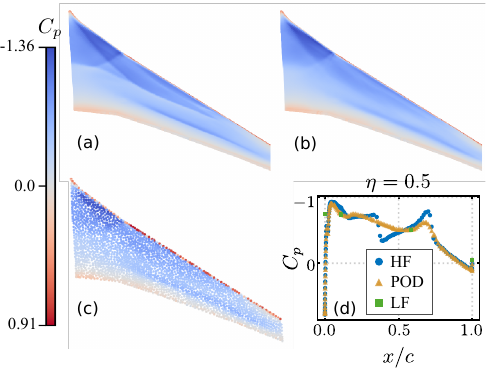}
    \caption{POD + FPS fidelity reduction: $C_p$ vectors for (a) original HF, (b) truncated POD, and (c) LF fields. (d) Details of the data representation.}
    \label{down_POD_FPS}
\end{figure}

Mathematically, a filtering is applied to this matrix via Singular Value Decomposition (SVD):
\begin{equation}
\mathbf{X} = \mathbf{U}\,\boldsymbol{\Sigma}\,\mathbf{V}^T
= \sum_{i=1}^{r} \sigma_i \,\mathbf{u}_i \mathbf{v}_i^T .
\end{equation}

where $\mathbf{U}$ contains the spatial modes, $\boldsymbol{\Sigma}$ is the diagonal matrix of singular values $\sigma_i$, and $\mathbf{V}$ represents the temporal coefficients.

To simulate the deficit in physical fidelity, a modal truncation retaining only the first $r^*$ modes is applied. This truncation acts as an \textit{information retained} filter, where the degree of fidelity reduction is controlled directly by the cumulative relative energy. The truncation threshold is determined by:
\begin{equation}
E_{r^*} = \frac{\sum_{i=1}^{r^*} \sigma_i^2}{\sum_{i=1}^r \sigma_i^2}.
\end{equation}

The reconstructed field $\mathbf{X}_{r^*}$, which retains only the large-scale coherent structures, is given by:
\begin{equation}
\mathbf{X}_{r^*} = \mathbf{U}_{r^*}\,\boldsymbol{\Sigma}_{r^*}\,\mathbf{V}_{r^*}^T .
\end{equation}

Physically, a lower energy retention implies using fewer modes, which preserves the dominant features but fails to reconstruct fine details and high-frequency structures. In the context of aerodynamic data, this truncation causes a notable blurring of sharp nonlinearities, such as shock waves (\autoref{down_POD_FPS}), thereby simulating a numerical model of lower precision or higher numerical dissipation.

Additionally, although not employed in the final results of this paper, the implementation allows for further signal degradation. Depending on the typology of the source data, Gaussian noise ($\mathcal{N}(0, \sigma_{noise}^2)$) or a constant bias can be superimposed onto the reconstructed data to simulate sensor noise or systematic model errors.

\subsubsection{Resolution Reduction via FPS}
Once the precision-reduced dataset $\mathbf{X}_{r^*}$ has been generated, the next step is to reduce the number of available data points. This is a fundamental step for evaluating the resolution enhancement capability of the autoencoder.

To this end, the Farthest Point Sampling (FPS) algorithm was employed, as it reduces point cloud density while preserving the overall geometric distribution of the data \cite{FPS_art}. It is important to note that FPS operates strictly within the physical space (i.e., the spatial coordinates of the wing mesh) rather than the pressure coefficient data space. FPS is an iterative greedy sampling strategy that seeks to maximize the spatial coverage of the selected subset. Let $P$ denote the set of physical coordinate vectors already selected. At each iteration, the next spatial point $\mathbf{p}_*$ is chosen from the remaining grid locations based on its physical distance to $P$ according to the Max-Min criterion:
\begin{equation}
\mathbf{p}_* = \operatorname*{argmax}_{\mathbf{p}\notin P} \left( \min_{\mathbf{q}\in P} \lVert \mathbf{p}-\mathbf{q} \rVert_2 \right).
\end{equation}

The algorithm is initialized with a randomly selected spatial coordinate and continues until the sampled subset reaches the desired cardinality, $|P| = D_{LF}$. By repeatedly selecting the point farthest from the current subset, FPS promotes a near-uniform spatial distribution of samples across the domain. Once the geometric subset $P$ is finalized, the corresponding rows are extracted from $\mathbf{X}_{r^*}$. The resulting dataset constitutes the final low-fidelity representation, combining both reduced numerical precision and reduced spatial resolution.

\subsection{Reduction via FPS and k-Nearest Neighbors (KNN) Averaging}
\begin{figure}[H]
    \centering
    \includegraphics[width = 8.2 cm]{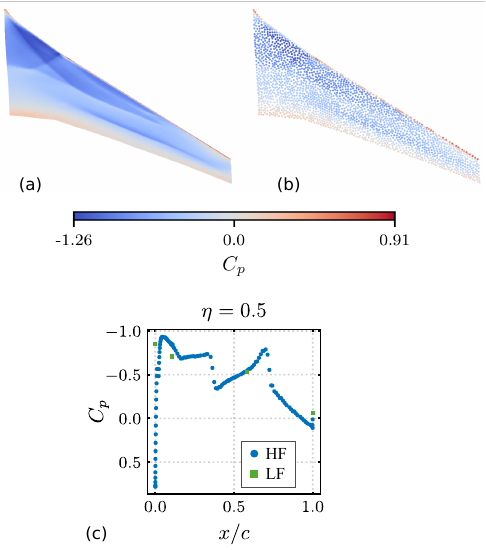}
    \caption{FPS + KNN fidelity reduction: $C_p$ vectors for (a) original HF, and (b) LF fields. (d) Details of the data representation.}
    \label{down_FPS}
\end{figure}

A second implemented down-fidelity method combines FPS with a local averaging technique based on the $k$-Nearest Neighbors (KNN) algorithm. In contrast to the previous technique, which first degraded the precision of the entire field and subsequently downsampled, this approach performs both operations simultaneously (\autoref{down_FPS}). The objective is to generate a low-resolution mesh where the value at each point represents a local mean of the high-fidelity field.

Data reduction is achieved through a hybrid FPS-KNN approach. Initially, the mesh geometry is constructed by selecting $D_{LF}$ seed points from the HF set using the FPS maximum distance metric. In the second stage, field reconstruction is performed: instead of direct pointwise assignment, the value at each LF node is computed as the arithmetic mean of its $k$ nearest neighbors in the original $D_{HF}$ cloud.

Thus, both fidelity reduction objectives are met. First, the resolution is drastically reduced by transitioning from $D_{HF}$ points to $D_{LF}$ points. Second, local precision is compromised by the averaging process. This averaging process acts as a low-pass filter, blurring the sharp gradients and high-frequency features present in the HF data. The result simulates a lower fidelity model that captures the average behavior within a region, rather than the exact pointwise value at that coordinate.

\subsection{Voxelization}
\begin{figure}[H]
    \centering
    \includegraphics{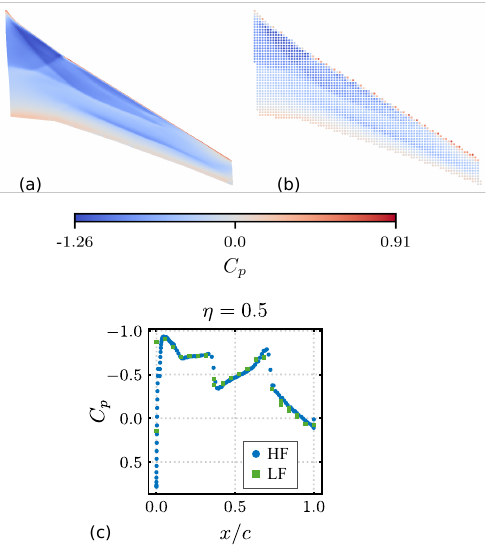}
    \caption{Voxelization fidelity reduction: $C_p$ vectors for (a) original HF, and (b) LF fields. (d) Details of the data representation.}
    \label{down_vox}
\end{figure}

The voxelization method consists of the discretization of a three-dimensional spatial domain into a regular grid of volumetric elements \cite{Kaufman1996}, known as voxels (\autoref{down_vox}). When applied to a point cloud, the algorithm superimposes a 3D mesh over the HF dataset, where the size of each mesh cell is defined by a voxel size parameter.

The fidelity reduction and resolution occur simultaneously during this process. The algorithm first identifies all HF points that reside within the boundaries of each individual voxel. For each voxel containing at least one HF point, a single LF point is computed.

Resolution reduction is direct, as it collapses a large number of HF points into a much smaller set of occupied voxels. The new LF dataset is composed of the coordinates of the centers of these occupied voxels. The property value assigned to each LF voxel center is not a measured value from the HF set, but rather the arithmetic mean of the property values of all HF points falling within that voxel. Similar to the KNN averaging described in Section~B, this process acts as a spatial low-pass filter, removing high-frequency details and blurring local gradients, thereby simulating a lower-resolution measurement or simulation.

Additionally, an optimization has been implemented for complex geometries. Optionally, prior to voxelization, the point cloud can be rotated using Principal Component Analysis (PCA) to align it with its principal axes. This ensures a more efficient discretization for elongated or non-axis-aligned geometries. Upon computing the voxel centers, the inverse rotation transformation is applied to map the coordinates back to the original reference frame.

\subsection{Uniform Quantization}
\begin{figure}[H]
    \centering
    \includegraphics{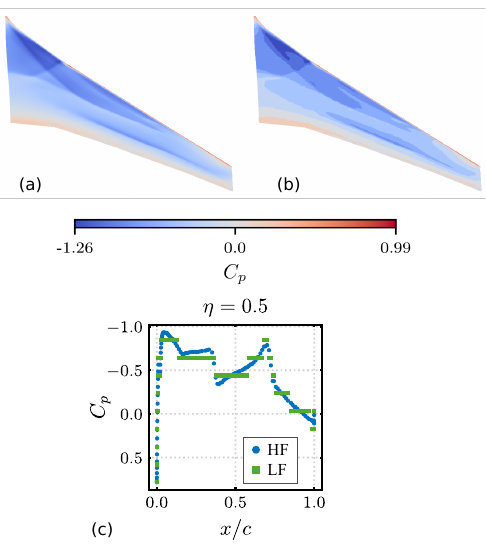}
    \caption{Uniform Quantization fidelity reduction: $C_p$ vectors for (a) original HF, and (b) LF fields. (d) Details of the data representation.}
    \label{down_quant}
\end{figure}

Quantization is a process that reduces data fidelity by decreasing numerical precision \cite{Widrow1996}. The method involves mapping a range of high-resolution values to a finite, discrete set of output values, known as quantization levels (\autoref{down_quant}).

In this implementation, uniform quantization is applied. The algorithm first identifies the global minimum and maximum values across the entire HF dataset. This range $[\text{min}, \text{max}]$ is then uniformly divided into a predefined number of levels. Subsequently, each individual value in the HF data is rounded or assigned to the nearest discrete level.

Loss of fidelity arises from the discretization of continuous value space. Fine details and subtle variations in the data, which originally lay between two quantization levels, are reduced to the same value. This process introduces an error analogous to reducing bit depth or using a measuring instrument with lower resolution.

It is important to note that this method, by itself, does not alter the data resolution or dimensionality. The resulting quantized data matrix retains the exact number of points as the HF input matrix. To achieve a reduction in both precision and resolution, the quantization step must be followed by a downsampling method, such as the FPS, to select a subset of the already quantized points.
\end{document}